\documentclass{article}

\usepackage{algorithm}
\usepackage{algorithmic}

\usepackage{microtype}
\usepackage{graphicx}
\usepackage{subfig}
\usepackage{stackengine}
\usepackage{booktabs}
\usepackage{hyperref}

\usepackage{authblk}
\usepackage{fullpage}

\usepackage{amsthm}
\usepackage{mathrsfs}
\usepackage{mathtools}
\usepackage{amsmath}
\usepackage{amssymb}
\usepackage{cancel}
\usepackage{breqn}
\usepackage{enumitem}
\usepackage{Definitions}

\newcommand{\xhdr}[1]{\vspace{1mm} \noindent {\bf #1.}}
\newcommand{\given}{\,|\,}
\newcommand{\lgiven}{\,\bigg\rvert\,}
 
\newcommand*{\range}{\mathrm{Range}}

\newcommand{\algnameNS}{Calibrated Subset Selection}
\newcommand{\algname}{\algnameNS\ }
\newcommand{\algabbrNS}{CSS}
\newcommand{\algabbr}{\algabbrNS\ }
\newcommand{\algdivabbrNS}{\algabbr (Diversity)}
\newcommand{\algdivabbr}{\algdivabbrNS\ }

\title{Improving Screening Processes via Calibrated Subset Selection}

\author[1]{Lequn Wang\thanks{lw633@cornell.edu. Most of the work was done during Wang'{}s internship at the Max Planck Institute for Software Systems.}}
\author[1]{Thorsten Joachims\thanks{tj@cs.cornell.edu}}
\author[2]{Manuel Gomez Rodriguez\thanks{manuelgr@mpi-sws.org}}

\affil[1]{Department of Computer Science, Cornell University}
\affil[2]{Max Planck Institute for Software Systems}

\date{}

\begin{document}

\maketitle

\begin{abstract}
Many selection processes such as finding patients qualifying for a medical trial or retrieval pipelines in search 
engines consist of multiple stages, where an initial screening stage focuses the resources on shortlisting the most 
promising candidates. 
In this paper, we investigate what guarantees a screening classifier can provide, independently of whether it is 
constructed manually or trained. 
We find that current solutions do not enjoy distribution-free theoretical guarantees---we show that, in general, 
even for a perfectly calibrated classifier, there always exist specific pools of candidates for which its shortlist 
is suboptimal. 
Then, we develop a distribution-free screening algorithm---called \algname (\algabbrNS)---that, given any classifier and some amount of calibration 
data, finds near-optimal shortlists of candidates that contain a desired number of qualified candidates in expectation. 
Moreover, we show that a variant of CSS that calibrates a given classifier multiple times across specific 
groups can create shortlists with provable diversity gua\-ran\-tees. 
Experiments on US Census survey data validate our theoretical results and show that the shortlists provided by our 
algorithm are superior to those provided by several competitive baselines.

\end{abstract}

\vspace{-1mm}
\section{Introduction}
\label{sec:introduction}
\vspace{-1mm}
Screening is an essential part of many selection processes, where an often intractable number of candidates is reduced to a shortlist of the most promising candidates for detailed---and more resource intensive---evaluation. Screening thus enables an allocation of resources that improves the overall quality of the decisions under limited resources. Examples of such screening problems are: finding patients in a large database of electronic health records to manually evaluate for qualification to take part in a medical trial~\cite{liu2021evaluating}; the first stage of a multi-stage retrieval pipeline of a search engine~\cite{covington2016deep,geyik2019fairness}; or which people to reach out to with a personalized invitation to apply to a specific job posting~\cite{raghavan2020mitigating}.  
In each of these examples, there is significant pressure to make high-qua\-li\-ty, 
unbiased screening decisions quickly and efficiently, often about thousands or even millions 
of candidates under limited re\-sour\-ces and additional diversity requirements~\cite{bendick1997employment,bertrand2004emily,johnson2016if,covington2016deep}.
While these screening decisions have been made manually or through manually constructed rules in the past, automated predictive tools for optimizing screening decisions are becoming more prevalent~\cite{cowgill2018bias, chamorro2019should, raghavan2020mitigating, sanchez2020does}.

Algorithmic screening has been typically studied together with other high-stakes decision making problems as a
supervised learning problem~\cite{corbett2017algorithmic, kilbertus2020fair,sahoo2021reliable}. Under this perspective, algorithmic screening reduces to: 
%
(i) training a classi\-fier that estimates the probability that 
a candidate is \emph{qualified} given a set of observable features;
%
(ii) de\-signing a deterministic threshold rule that shortlists candidates by 
thresholding the candidates'{} pro\-ba\-bi\-li\-ty values estimated by the 
classifier.
%
Here, the classifier and the threshold rule aim to maximize a measure of average accuracy 
and average utility, respectively, possibly subject to diversity constraints.
Only very recently, it has been argued that the classifier must also satisfy threshold 
calibration, a specific type of group calibration, to be able to accurately estimate the
average utility of the threshold rule~\cite{sahoo2021reliable}.

Unfortunately, the above screening algorithms do not enjoy dis\-tri\-bu\-tion-free guarantees\footnote{We refer to distribution-free guarantees as finite-sample distribution-free guarantees, since we never have infinite amount of data in practice. } on the quality of the shortlisted candidates. 
As a result, they may not always hold their promise of increasing the 
efficiency of a selection process without decreasing the quality of the screening decisions.
The results in this paper bridge this gap. 
In particular, we focus on developing screening algorithms that, without making any distributional assumptions on the candidates, provide the smallest shortlists of candidates, among those in a given pool, con\-tai\-ning a desired expected number of qualified candidates with high probability.\footnote{If the overall pool of candidates does not contain the desired number of qualified candidates with high probability, the algorithms should also determine that.}


\xhdr{Our Contributions}
%
%
We first show that, even if a classifier is perfectly calibrated, in general,
there always exist\- specific pools of candidates for which the shortlist created 
by a policy that makes decisions based on the probability predictions from the classifier 
will be significantly suboptimal, both in terms of the size and the expected number of 
qualified candidates.
Then, we develop a distribution-free screening algorithm---called \algname (\algabbrNS)---that cali\-brates any given 
classifier using calibration data in a way such that the shortlist of candidates 
created by thresholding the candidates'{} probability values estimated by the calibrated 
classifier is near-optimal in terms of expected size, and it
provably contains, in expectation across all potential pools of candidates, 
a desired number of qualified candidates with high probability.
Moreover, we theoretically characterize how the accuracy of the classifier and the amount of 
calibration data affect the expected size of the shortlists \algabbr provides. 
In addition, motivated by the Rooney rule~\cite{collins2007tackling}, which requires that,
when hiring for a given position, at least one candidate from the underrepresented group 
be interviewed, 
we demonstrate that a variant of \algabbr that calibrates the given classifier
multiple times across different groups can be used to create a shortlist with provable 
diversity guarantees---this shortlist contains, in expectation across all potential 
pools of candidates, a desired number of qualified candidates from 
each group with high probability.

Finally, we validate \algabbr on simulated screening processes created using  US Census survey 
data~\cite{ding2021retiring}\footnote{Our code is accessible at \url{ https://github.com/LequnWang/Improve-Screening-via-Calibrated-Subset-Selection.} }.
The results show that, compared to several competitive baselines, \algabbr consistently 
selects the shortest shortlists of candidates among those methods that select enough qualified candidates.
Moreover, the results also demonstrate that the amount of human effort \algabbr helps reducing---the 
difference in size between the pools of candidates and the shortlists it provides---depends on the accuracy 
of the classifier it relies upon as well as the amount of calibration data. 
However, the expected quality of the shortlists---the expected number of qualified candidates 
in the shortlists---never decreases below the user-specified requirement.
%

\xhdr{Further Related Work}
Our work builds upon prior literature on distribution-free uncertainty quantification, which includes calibration and conformal prediction. 

Calibration~\cite{brier1950verification,dawid1982well,platt1999probabilistic,zadrozny2001obtaining,gneiting2007probabilistic,gupta2020distribution} measures the accuracy of the probability outputs from predictive models. Many notions of calibration have been proposed to measure the accuracy of the probability outputs from a classifier~\cite{platt1999probabilistic,guo2017calibration} or a regression model~\cite{gneiting2007probabilistic,kuleshov2018accurate}. Arguably, the most commonly used notion is marginal (or average) calibration~\cite{gupta2020distribution,zhao2020individual,gneiting2007probabilistic,kuleshov2018accurate} for both classifiers and regression models, which requires the probability outputs be marginally calibrated on the whole population. 
Our definition of perfectly calibrated classifier inherits from the marginal calibration definition for classifiers in prior works~\cite{gupta2020distribution}. We also show how \algabbr produces an approximately calibrated regression model under average calibration definition in regression~\cite{gneiting2007probabilistic,kuleshov2018accurate}. In the other extreme, individual calibration~\cite{zhao2020individual} refers to a classifier that predicts the probability distribution for each example. We refer to classifiers with such properties the omniscient classifiers in this paper. 
Many recent works~\cite{chouldechova2017fair,kleinberg2016inherent,pleiss2017neurips} discuss the relationship between calibration and fairness in binary classification, and show that marginal calibration is incompatible with many fairness definitions in machine learning. Motivated by multicalibration~\cite{hebert2018multicalibration,jung2021moment}, which requires the predictive models be calibrated on multiple groups of candidates, we propose a variant of \algabbr that selects calibrated subsets within each group to ensure diversity in the final selected shortlists. Very recently, some works~\cite{sahoo2021reliable,zhao2021calibrating,straitouri2022provably,wang2022fairness} realize that we can design calibration algorithms (and calibration definitions) well-suited for different downstream tasks to achieve better performance. \algabbr is specifically designed for screening processes so that it provides near-optimal shortlists in terms of the expected size among those having distribution-free guarantees on the expected number of qualified candidates.

Conformal prediction~\cite{vovk1999machine,vovk2005algorithmic,shafer2008tutorial,romano2019conformalized,balasubramanian2014conformal, gupta2020distribution, angelopoulos2021gentle,chzhen2021set} aims to build confidence intervals on the probability outputs from predictive models. Within this literature, the work most closely related to ours is arguably the work by \cite{bates2021distribution}, which 
has focused on generating set-valued predictions from a black-box predictor that controls 
the expected loss on future test points at a user-specified level.
While one can view our problem from the perspective of set-valued predictions, applying 
their me\-tho\-do\-lo\-gy to find near-optimal solutions in our problem is not straightforward, and one would need 
to assume to have access to qualification labels for all the candidates in multiple pools, something 
we view as rather impractical. 

Moreover, our work also builds on work in budget-constrained decision making, where one needs to first select a
set of candidates to screen and then, given the result of that screening process, determine to whom to allocate
the resources~\cite{cai2020fair, bakker2021beyond}. 
This contrasts with our work where all candidates undergo screening. 

Many large-scale recommender systems adopt multi-stage pipelines~\cite{bendick2013developing}. Existing works on
multi-stage recommender systems~\cite{ma2020off,hron2021component} focus on learning accurate classifiers in
different stages. Complementary to these works, we assume the classifiers are already given and provide
distribution-free and finite-sample guarantees on the quality of the shortlists. 

\vspace{-1mm}
\section{Problem Formulation}
\label{sec:problem-formulation}
\vspace{-1mm}
Given a candidate with a feature vector $x \in \Xcal$, 
we assume the candidate can be either qualified 
($y = 1$) or unqualified ($y = 0$) for the selection objective\footnote{In practice, one needs to measure 
qualification using proxy variables and these proxy variables need to be chosen carefully to not perpetuate 
historical biases~\cite{bogen2018help,garr2019diversity,tambe2019artificial}.}. 
Moreover, let $f \,:\, \Xcal \rightarrow [0, 1]$ be a classifier that maps a candidate'{}s feature vector $x$ 
to a quality score\footnote{Our theory and algorithms apply to any classifier with a bounded range, by scaling the scores to $[0,1]$.} $f(x)$. The higher the quality score $f(x)$, the more the classifier believes 
the candidate is qualified. 
%
%
%
Given a pool of $m$ candidates with feature vectors $\xb~=~\{x_i\}_{i \in [m]}$, 
an algorithmic screening policy \- $\pi \,:\, [0, 1]^{m} \rightarrow \Pcal(\{0, 1\}^{m})$ maps the candidates'{} quality scores 
$\{f(x_i)\}_{i \in [m]}$ to a pro\-ba\-bi\-li\-ty distribution over shortlisting 
decisions $\sbb = \{s_i\}_{i \in [m]}$.
Here, each shortlisting decision $s_i$ specifies whether the corresponding candidate is 
shortlisted ($s_i = 1$) or is not shortlisted ($s_i = 0$).
For brevity, we will write $\Sbb \sim \pi$ whenever there is no ambiguity\footnote{We use upper case letters to denote random variables, and lower case letters to denote realizations of random variables.}. 
%
Furthermore, we use $\pi_f$ to indicate that a policy makes shortlisting decisions based on the quality scores 
from classifier $f$. 
%
%
This implies that for any pool of candidates $\xb$ and any two candidates $i,j\in[m]$ in the pool, if $f(x_i)=f(x_j)$, then 
$\Pr(S_i=1) = \Pr(S_j=1)$.
We denote the set of all possible policies $\pi_f$ as $\Pi_f$. 

For any screening process, we would ideally like a scree\-ning policy $\pi$ that shortlists \emph{only} candidates who are qualified ($y = 1$).
Unfortunately, as long as there is no deterministic mapping between $x$ and $y$, such a \emph{perfect} 
screening policy $\pi$ does not exist in general.
Instead, our goal is to find a screening policy $\pi$ that shortlists a small set of candidates that provably contains \emph{enough} qualified candidates, without making any assumptions about the data distribution. 
%
These shortlisted candidates will then move forward in the selection process and will be evaluated in detail, possibly multiple times, until one or more qualified candidates are selected. 

More specifically, 
let each candidate'{}s feature vector $x$ and label $y$ be sampled from a data distribution $P_{X,Y} = P_{X} \times P_{Y \given X}$. 
Then, for a pool of $m$ candidates\footnote{For ease of presentation, we assume a constant pool size $m$ in 
the main paper. However, all the theoretical results and algorithms can be easily adapted to settings where 
the pool size changes across selection processes. Refer to Appendix~\ref{app:dynamic-pool-size} for more 
details.} with feature vectors $\Xb = \{X_i\}_{i\in[m]}$ and unobserved labels $\Yb = \{Y_i\}_{i\in[m]}$, 
where $X_i \sim P_X$, $Y_i\sim P_{Y\mid X_i}$ for all $i\in[m]$, we will investigate to what extent it is 
possible to find screening policies $\pi$ with near-optimal guarantees with respect to two different oracle 
policies. 
In particular: 
\begin{itemize}[nolistsep, leftmargin=0.8cm]
\item[(i)] an oracle policy $\pi^{\star}$ that, for any set of candidates $\xb\in\Xcal^m$, shortlists the smallest set of candidates that contains, in expectation with respect to $P_{Y \given X}$, more than $k$ qualified candidates, \ie, $\forall \xb\in\Xcal^m$, 
\begin{equation}
    \pi^{\star} \in \argmin_{\pi \in \Pi} \, \EE_{\Sbb \sim \pi}\left[ \sum_{i \in [m]} S_i \right],
    \label{eq:oracle-policy-individual}
\end{equation}
where 
\begin{equation*}
\Pi = \left\{ \pi \lgiven \EE_{\Sbb \sim \pi, \Yb \sim P} \left[ \sum_{i \in [m]} S_i Y_i \right] \geq k \right\}.
\end{equation*}
%
\item[(ii)] an oracle policy $\pi^{\star\star}$ that shortlists the smallest set of candidates that contains, in expectation with respect to $P_{X,Y}$, more than $k$ qualified candidates, \ie, 
\begin{equation}
       \pi^{\star\star} \in \argmin_{\pi \in \Pi} \, \EE_{\Xb \sim P, \Sbb \sim \pi} \left[ \sum_{i \in [m]} S_i \right], 
    \label{eq:oracle-policy-marginal}
\end{equation}
where 
\begin{equation*}
\Pi = \left\{ \pi \lgiven \EE_{(\Xb, \Yb) \sim P, \Sbb \sim \pi} \left[ \sum_{i \in [m]} S_i Y_i \right] \geq k \right\}.
\end{equation*}
%
%
\end{itemize}
In the language of distribution-free uncertainty quantification~\cite{balasubramanian2014conformal}, the optimality 
guarantees with respect to the first and the second oracle policy can be viewed as individual and marginal guarantees
respectively.

\vspace{-1mm}
\section{Impossibility of Screening with Individual Guarantees}
\label{sec:individual}
\vspace{-1mm}
If we had access to an omniscient classifier $f^{\star}(x) = \Pr(Y = 1 \given X = x)$ 
for all $x \in \Xcal$, then we could recover the oracle policy 
$\pi^{\star}$ defined by Eq.~\ref{eq:oracle-policy-individual} just 
by thres\-hol\-ding the quality scores of each candidate in the pool.
More specifically, we have the following theorem\footnote{All proofs 
can be found in Appendix~\ref{app:proofs}.}.
\begin{theorem} 
\label{thm:omniscient-classifier-individual}
%
The screening policy $\pi^\star_{f^{\star}}$ that, given any pool of $m$ candidates with feature vectors $\xb\in\Xcal^m$,  takes shortlisting decisions as
\begin{equation}
\label{eq:decision_rule_omniscient_classifier}
s_i =
\begin{cases}
1 & \textnormal{if} \, f^{\star}(x_i)> t^\star, \\
\textnormal{Bernoulli}(\theta^\star) & \textnormal{if} \, f^{\star}(x_i) = t^\star, \\
0 & \textnormal{otherwise},
\end{cases}
\end{equation}
where
\begin{equation*}
t^\star = \sup\left\{t\in[0,1] \lgiven \sum_{i\in[m]}\II\{f^{\star}(x_i)\geq t\}f^{\star}(x_i)\geq k \right\}
\end{equation*}
and
\begin{equation*}
    \theta^\star = \frac{k - \sum_{i\in[m]}\II\{f^{\star}(x_i)> t^\star\}f^{\star}(x_i)}{\sum_{i\in[m]}\II\{f^{\star}(x_i)= t^\star\}f^{\star}(x_i)},
\end{equation*}
is a solution (if there is one) to the constrained minimization problem  defined in Eq.~\ref{eq:oracle-policy-individual}. 
\end{theorem}
Unfortunately, without distributional assumptions about $P_{X,Y}$, finding the omniscient classifier $f^{\star}$ from 
data is impossible, even asymptotically, if the distribution $P_{f^{\star}(X)}$ induced by $P_{X,Y}$ is nonatomic, as recently shown by Barber~\cite{barber2020distribution} and Gupta et al.~\cite{gupta2020distribution}.
Alternatively, one may think whether there exist other classifiers $h$ allowing for near-optimal screening 
policies $\pi_h$ with individual gua\-ran\-tees. 
In this context, a natural alternative is a perfectly calibrated classifier $h$ other than the omniscient 
classifier $f^{\star}$. A classifier $h$ is perfectly calibrated if and only if
\begin{equation} \label{eq:perfectly-calibrated}
    \Pr(Y=1 \given h(X) = a) = a \quad \forall a\in \range(h).  
\end{equation}
However, the following theorem shows that there exist\- data distributions $P_{X,Y}$ for which
there is no perfectly calibrated classifier $h \neq f^{\star}$ allowing for a screening policy $\pi_h$ 
with individual guarantees of optimality.
\begin{proposition} \label{prop:impossibility-individual-guarantees}
Let $\Xcal = \{ a,b \}$, $\Pr(Y = 1 \given X =a) = 1$, $\Pr(Y = 1 \given X = b) = \frac{k}{m}$,
and $f^{\star}$ be the omniscient classifier.
Then, for any screening policy $\pi_h$ using any perfectly calibrated classifier $h \neq f^{\star}$, 
there exists a pool of candidates $\xb=\cbr{x_{i}}_{i\in[m]}\in\Xcal^m$ such that 
\begin{equation*}
\left|\EE_{\Sbb \sim \pi_h, \Sbb^{\star} \sim \pi^{\star}_{f^\star}} \left[\sum_{i\in[m]} \left(S_{i} - S^{\star}_{i}\right) \right] \right|\geq \frac{m-k}{2}; 
\end{equation*}
%
and a pool of candidates $\xb'=\cbr{x'_{i}}_{i\in[m]}\in\Xcal^m$ such that 
\begin{equation*}
\left|\EE_{\Yb \sim P, \Sbb \sim \pi_h, \Sbb^{\star} \sim \pi^{\star}_{f^\star} } \!\!\!\left[\!\sum_{i\in[m]} \!\!\!\left( S_{i} \!-\! S^{\star}_{i} \right)\! Y_i \right] \!\right|
\geq \frac{k}{2}\!\left(1\!-\!\frac{ k}{m}\right).
\end{equation*}
%
\end{proposition}
%
The above result reveals that, if $m\gg k$, for any screening policy $\pi_h$ using a calibrated classifier $h \neq f^{\star}$,  there are always scenarios in which $\pi_h$ provides shortlists that are $\frac{m}{2}$ apart from those provided by the oracle policy $\pi^{\star}$ in size or $\frac{k}{2}$ apart in terms of the number of qualified candidates. 
In particular, the second part of Proposition~\ref{prop:impossibility-individual-guarantees} directly implies that there is no policy $\pi\in\Pi_h$ that satisfies the individual guarantee. 
%
%
This negative result motivates us to look for screening policies with marginal guarantees of optimality. 

\vspace{-1mm}
\section{Screening Algorithms with Marginal Guarantees}
\label{sec:marginal}
\vspace{-1mm}
To investigate to what extent it is possible to find screening policies with near-optimal guarantees with respect to the oracle policy $\pi^{\star\star}$ defined in Eq.~\ref{eq:oracle-policy-marginal}, we focus on perfectly calibrated classifiers $h$ with finite range, \ie, $|\text{Range}(h)| < \infty$.
The reason is that, similarly as in the case of the omnis\-cient classifier $f^{\star}$, it is impossible to find nonatomic perfectly calibrated classifiers $h$ from data, even asymptotically~\cite{barber2020distribution,gupta2020distribution}. 
We will first introduce the optimal algorithm assuming we have access to a perfectly calibrated classifier, and 
discuss how the accuracy of the classifier affects the performance of the optimal algorithm. 
Then, we will introduce the proposed \algabbr algorithm that, given a classifier 
and some amount of calibration data, selects shortlists that are near-optimal in terms of the expected size, while ensuring that the expected number of qualified candidates in the shortlists is above the desired threshold. 

\xhdr{An Optimal Calibration Algorithm with Perfectly Calibrated Classifiers}
Let $h$ be a perfectly calibrated classi\-fier with $\text{Range}(h) = \{\mu_b\}_{b\in[B]}$, 
and denote $\rho_b = \EE_{X \sim P}[\II\{h(X) = \mu_b\}]$. 
First, note that the classifier $h$ induces a sample-space \emph{partition} of $\Xcal$
into $B$ regions or bins $\{\Xcal_b\}_{b \in [B]}$,
where $\mu_b = \Pr(Y = 1 \given X \in \Xcal_b)$ and $\rho_b = \Pr(X \in \Xcal_b)$.
Then, the following theorem shows that the optimal screening policy $\pi^{\star}_h$ that shortlists the smallest set of candidates within the set of policies $\Pi_h$, which contain, in expectation with respect to $P_{X,Y}$, more than $k$ qualified candidates, is given by a threshold decision rule.
\begin{theorem} \label{thm:perfectly-calibrated-classifier-marginal}
The screening policy $\pi^{\star}_{h}$ that takes shortlisting decisions as
\begin{equation*}
s_i =
\begin{cases}
1 & \textnormal{if} \, h(x_i)> t_h, \\
\textnormal{Bernoulli}(\theta_h) & \textnormal{if} \, h(x_i) = t_h, \\
0 & \textnormal{otherwise},
\end{cases}
\end{equation*}
where
\begin{equation*}
t_h = \sup\left\{\mu\in \{\mu_b\}_{b\in[B]} \lgiven \sum_{b\in[B]}\mu_{b}\II\{\mu_b\geq \mu\}\rho_b\geq \frac{k}{m}\right\}
\end{equation*}
and
\begin{equation*}
    \theta_h = \frac{k/m - \sum_{b\in[B]}\mu_{b}\II\{\mu_b> t_h\}\rho_b}{\sum_{b\in[B]}\mu_{b}\II\{\mu_b=t_h\}\rho_b},
\end{equation*}
is a solution (if there is one) to the constrained minimization problem defined in Eq.~\ref{eq:oracle-policy-marginal} over the set of policies $\Pi_h$. 
\end{theorem}
However, it is important to realize that the expected size of the shortlists provided by 
$\pi^{\star}_h$ for different perfectly calibrated classifiers $h$ may differ. 
To put it differently, not all screening policies $\pi^{\star}_h$ (and classifiers $h$) will help reduce the downstream effort by the same amount.
To characterize this diffe\-rence, we introduce a notion of dominance between perfectly calibrated classifiers.
\begin{definition}
Let $h$ and $h'$ be perfectly calibrated classifiers.
We say that $h$ dominates $h'$ if for any $x_1, x_2 \in \Xcal$ such that $h(x_1) = h(x_2)$, it holds that 
$h'(x_1) = h'(x_2)$.
\end{definition}
Equipped with this notion, we now characterize the diffe\-rence in size between shortlists provided by different perfectly calibrated classifiers using the following corollary, which follows  
from Theorem~\ref{thm:perfectly-calibrated-classifier-marginal}.
\begin{corollary} \label{cor:dominance}
Let $h$ and $h'$ be perfectly calibrated classifiers.
If $h$ dominates $h'$, then
\begin{equation*}
\EE_{\Xb \sim P, \Sbb \sim \pi^\star_{h}, \Sbb' \sim \pi^\star_{h'}}\left[\sum_{i\in[m]} \rbr{S_i - S'_i} \right] \leq 0. 
\end{equation*}
\end{corollary}
This notion of dominance relates to the notion of sharpness~\cite{gneiting2007probabilistic,kuleshov2018accurate}, which links the accuracy of a calibrated classifier to how fine-grained the calibration is within the sample-space. In particular, if $h$ dominates $h'$, it can be readily shown that $h$ is sharper than $h'$. However, existing works have not studied the effect of the sharpness of a classifier on its performance on screening tasks.

\xhdr{A Near-Optimal Screening Algorithm with Calibration Data}
Until here, we have assumed that a perfectly calibrated classifier $h$ with finite range, as well as the size 
of each bin $\rho_b$ are given. 
However, using finite amounts of calibration data, we can only hope to find \emph{appro\-xi\-ma\-te\-ly} calibrated classifiers.  
%
Next, we will develop an algorithm that, rather than training an approximately calibrated classifier from scratch, it \emph{approximately} calibrates a given classi\-fier $f$, \eg, a deep neural network, using a calibration set 
$\Dcal_{\text{cal}} = \{ (x_i^c, y_i^c) \}_{i \in [n]}$, 
which are independently sampled from $P_{X,Y}$\footnote{Superscript $c$ is used to differentiate between candidates in the calibration set and candidates in the pool at test time.}.
%
%
In doing so, the algorithm will find the optimal sample-space partition and decision rule that minimize the expected size of the provided shortlists among those ensuring that 
an empirical lower bound on the expected number of qualified candidates is greater 
than $k$.

From now on, we will assume that the given classifier $f$ is nonatomic\footnote{We can add arbitrarily small noise to break atoms. } and sa\-tis\-fies a natural 
monotonicity property\footnote{The monotonicity property we consider is different 
from that considered in the literature on monotonic classification~\cite{cano2019monotonic}.} with respect to the data distribution $P_{X,Y}$, a significantly weaker assumption than 
calibration.
\begin{definition}
A classifier $f$ is monotone with respect to a data distribution $P_{X,Y}$ if, for any $a, b\in\range(f)$ such that $a<b$, it holds
that
\begin{equation*}
\Pr\cbr{Y = 1 \given f(X) = a} \leq \Pr\cbr{Y = 1 \given f(X) = b}. 
\end{equation*}
\end{definition}
Under this assumption\footnote{
%
There is empirical evidence that well-performing classifiers learned from data are (approximately) monotone~\cite{guo2017calibration,kuleshov2018accurate,gupta2020distribution}.
In this context, we would like to emphasize that the monotone property 
is only necessary to prove that the expected size of the shortlists provided by our algorithm is near-optimal.
However, the distribution-free guarantees on the expected number of qualified shortlisted candidates holds even for non-monotone 
classifiers.
}, we can first show that 
the solution (if there is one) to the constrained minimization problem in Eq.~\ref{eq:oracle-policy-marginal} over $\Pi_f$ is a threshold decision rule $\pi_{f, t_f^{\star}}$ with some threshold $t_f^{\star}\in[0,1]$ that takes shortlisting decisions as
\begin{equation} \label{eq:decision-threshold-rule-algorithm}
s_i =
\begin{cases}
1 & \textnormal{if} \, f(x_i)\geq t_f^{\star}, \\
0 & \textnormal{otherwise}.
\end{cases}
\end{equation}
More specifically, we have the following 
theorem.
\begin{theorem}
\label{thm:optimality_threshold_decision_rules}
Let $f$ be a monotone classifier with respect to $P_{X,Y}$ and the 
distribution $P_{f(X)}$ induced by $P_{X,Y}$ is nonatomic. 
Then, for any $\pi_{f} \in \Pi_{f}$, there always exists a threshold decision 
rule $\pi_{f, t} \in \Pi_f$ with some threshold $t\in[0,1]$ such that 
\begin{equation*}
\EE_{\Xb\sim P, \Sbb \sim \pi_f, \Sbb' \sim\pi_{f, t}} \left[\sum_{i\in[m]} (S_i - S'_i) \right] \geq 0
\end{equation*}
and
\begin{equation*}
\EE_{(\Xb, \Yb) \sim P, \Sbb \sim \pi_f, \Sbb' \sim\pi_{f, t}}\left[\sum_{i\in[m]} Y_i (S_i - S'_i)\right] \leq 0. 
\end{equation*}
\end{theorem}
The theorem directly implies there exists a threshold decision policy $\pi_{f,t^\star_f}$ that is optimal in the constraint minimization problem in~\ref{eq:oracle-policy-marginal} (if there is a solution). 



As a result, we focus our attention on finding a near-optimal threshold decision policy. We first notice that each threshold decision policy $\pi_{f,t}$ induces a sample space partition of $\Xcal$ into two bins $\Xcal_{t,1} = \{ x \in \Xcal \given f(x) \geq t \}$ and $\Xcal_{t,2} = \{ x \in \Xcal \given f(x) < t \}$. Thus, it is sufficient to analyze the calibration errors  of the family of approximately calibrated classifiers $h_t$ with the two bins.  
In Appendix~\ref{app:screening-multiple-bins}, we show that using the calibration errors of calibrated classifiers that partition the sample-space into more bins will only worsen the performance guarantees of threshold policies. 
At this point, one may think of applying conventional dis\-tri\-bu\-tion-free calibration 
methods to bound the calibration errors of each classifier $h_t$ with high probability.
However, prior distribution-free calibration methods can only bound calibration errors on the mean values 
$\mu_b$ for each bin $b$, but not the bin sizes $\rho_b$. 
To make things worse, to find the threshold value $\hat{t}_f$ with optimal guarantees, we need to bound the 
calibration errors of all classifiers $h_t$, simultaneously with high pro\-ba\-bi\-li\-ty. 
Unfortunately, since $t \in [0, 1]$, there are infinitely many of them and we cannot naively apply a union 
bound on the bounds derived separately for each classifier.

To overcome the above issues, we will now leverage the Dvo\-retz\-ky–Kiefer–Wolfowitz–Massart (DKWM) inequality~\cite{dvoretzky1956asymptotic, massart1990tight}, 
which bounds how close an empirical cumulative distribution function (CDF) is to the cumulative distribution function of the distribution from which the empirical samples 
are sampled.
More specifically, let $\delta_{t, 1} := \EE_{(X,Y)\sim P}[\II\left\{f(X)\geq t\right\} Y]$ and 
\begin{equation*}
    \hat{\delta}_{t, 1} = \frac{1}{n} \sum_{i \in [n]} \II\left\{f(x_i^c)\geq t\right\} y^c_i
\end{equation*}
%
%
be an empirical estimator of $\delta_{t,1}$ using samples from the calibration set $\Dcal_{\text{cal}}$. 
Then, we can use the DKWM inequality to bound the calibration errors $|\hat{\delta}_{t, 1}-\delta_{t, 1}|$ across all approximately calibrated classifiers $h_t$ with high probability:
\begin{proposition}
\label{prop:bounds_errors_calibration}
For any $\alpha\in (0,1)$, with probability at least $1-\alpha$ (in $f$ and $\Dcal_{\textnormal{cal}}$), it holds that
\begin{equation*}
\left|\delta_{t,1}-\hat{\delta}_{t,1}\right|\leq\sqrt{\ln\left(2/\alpha\right)/(2n)}\coloneqq\epsilon(\alpha,n)  
\end{equation*}
simultaneously for all $t \in [0,1]$.
\end{proposition}
\begin{algorithm}[t]
\begin{algorithmic}[1]
\STATE{{\bf input:} $k$, $m$, $\Dcal_{\textnormal{cal}}$, $f$, $\alpha$, $\xb$}
\STATE{{\bf initialize:} $\sbb=\mathbf{0}$ }
\STATE{$\hat{t}_f=\sup\cbr{t\in[0,1]\lgiven \hat{\delta}_{t,1} \geq k/m + \epsilon(\alpha,n)}$}
\FOR{ $i=1,\dots,m$ }
\IF{$f(x_i)\geq \hat{t}_f$}
\STATE $s_i = 1$
\ENDIF
\ENDFOR
\STATE{{\bf return} $\sbb$}
\end{algorithmic}
\caption{\algname (\algabbrNS)}
\label{alg:screening}
\end{algorithm}
In Appendix~\ref{app:error-bounds-regression-calibration}, we show that based on the above error guarantees, we can build a regression model that achieves average calibration in regression~\cite{gneiting2007probabilistic,kuleshov2018accurate}, if we regard the binary classification problem as a regression problem. 
Further, building on the above proposition, we can derive an em\-pi\-ri\-cal lower bound on the expected number of qualified candidates in the shortlists provided by all the threshold decision rules $\pi_{f, t}$.
\begin{corollary} \label{cor:bounds_on_objectives}
For any $\alpha\in(0,1)$, with probability at least $1-\alpha$ (in $f$ and $\Dcal_{\textnormal{cal}}$), it holds that 
\begin{equation*}
\EE_{(\Xb,\Yb) \sim P, \Sbb\sim\pi_{f,t}}\left[\sum_{i\in[m]}Y_iS_i\right]
\geq m\rbr{\hat{\delta}_{t,1} - \epsilon(\alpha,n)} 
\end{equation*}
simultaneously for all $t \in [0,1]$.
\end{corollary}
Now, to find the decision threshold rule $\pi_{f, \hat{t}_f}$ that provides, in expectation, 
the smallest shortlists of candidates subject to a constraint on the lower bound on 
the expected number of qualified candidates in the provided shortlists, \ie,
\begin{equation} \label{eq:optimization_problem_calibration_from_data}
\hat{t}_f = \argmin_{t \in \Tcal} \, \EE_{\Xb\sim P, \Sbb\sim \pi_{f,t}}\left[\sum_{i\in[m]}S_i\right]
\end{equation}
where $\Tcal = \{ t\in[0,1] \given m\rbr{\hat{\delta}_{t,1} - \epsilon(\alpha,n)}\geq k \}$,
we resort to the following theorem.
\begin{theorem}
\label{thm:calibration_from_data_threshold}
The threshold value
\begin{equation}
\label{eq:calibration_from_data_threshold}
\begin{split}
\hat{t}_f = \sup\bigg\{t\in[0,1]\lgiven&\hat{\delta}_{t,1} \geq k/m + \epsilon(\alpha,n)\bigg\} 
\end{split}
\end{equation}
is a solution (if there is one) to the constrained minimization problem defined in Eq.~\ref{eq:optimization_problem_calibration_from_data}.
\end{theorem}
Algorithm~\ref{alg:screening} summarizes our resulting \algabbr algorithm, whose complexity
is $\Ocal(n\log (n))$. 

Finally, we can show that the expected size of the shortlists provided by \algabbr is near-optimal and we can also derive a lower bound on the worst-case size of the provided shortlists.
More specifically, the following propositions show that the difference in expected number of qualified
candidates between the shortlists provided by $\pi_{f, \hat{t}_f}$ and $\pi_{f, t^{\star}_f}$ decreases 
at a rate $1/\sqrt{n}$ and the worst-case size is on the order of $\sqrt{k}$.
\begin{proposition}
\label{prop:gap_calibration_from_data}
Let $f$ be a monotone classifier with respect to $P_{X,Y}$ and assume that the distribution $P_{f(X)}$ induced by $P_{X,Y}$ is nonatomic. Then, for any $\alpha\in(0,1)$, with probability at least $1-\alpha$, it holds that
\begin{equation*}
    \EE_{(\Xb,\Yb)\sim P, \Sbb \sim \pi_{f, t_f^{\star}}, \Sbb' \sim\pi_{f,\hat{t}_f}}\left[\sum_{i\in[m]} (S'_i - S_i) Y_i\right]
    \leq \frac{m}{n} + m\sqrt{2\ln(2/\alpha)/n}. 
\end{equation*}
\end{proposition}
\begin{proposition}
\label{prop:PAC_individual_guarantee}
For any $\alpha_1 \in(0,1),\alpha_2\in(e^{-\frac{9}{4}k},1)$ such that $\alpha_1 + \alpha_2<1$, \algabbr with parameter $\alpha = \alpha_1$ guarantees that, with probability at least $1-\alpha_1-\alpha_2$ (in $f$, $\Dcal_{\textnormal{cal}}$, and $\Xb,\Yb$), 
\begin{equation*}
\label{eq:worst-case-guarantee}
\sum_{i\in[m]}\!\!Y_iS_i\geq k - \frac{1}{3}\ln(1\!/\!\alpha_2)
- \frac{1}{3}\sqrt{\ln^2(1\!/\!\alpha_2) \!+\! 18 k\ln(1\!/\!\alpha_2)}. 
\end{equation*}
\end{proposition}
Finally, note that we can use the above worse-case guarantee to ensure that the worst-case size of the shortlists 
provided by \algabbr is greater than a target $t_{\textnormal{worst}}$ by setting $k$ to be 
slightly larger than $k_{\textnormal{worst}}$
\begin{equation*}
k = k_{\textnormal{worst}} + \frac{1}{3}\ln(1/\alpha_2) - \frac{1}{3}\sqrt{\ln^2(1/\alpha_2) + 18 k\ln(1/\alpha_2)}. 
\end{equation*}
%
%
By doing so, \algabbr will satisfy the constraints defined in Eq.~\ref{eq:oracle-policy-individual} 
with high probability with respect to the test pool of candidates.
%
%
However, \algabbr might not be optimal in terms of expected or worst-case shortlist size among those satisfying 
the above constrains.  
%
How to design screening algorithms that are optimal in terms of expected or worst-case shortlist size while sa\-tisfying
that each individual shortlist contains enough qualified candidates with high probability is an interesting problem 
to explore in future work.

\vspace{-1mm}
\section{Increasing the Diversity of Screening}
\label{sec:diversity}
\vspace{-1mm}
\begin{figure*}[t]
\centering
\includegraphics[width=1.\textwidth]{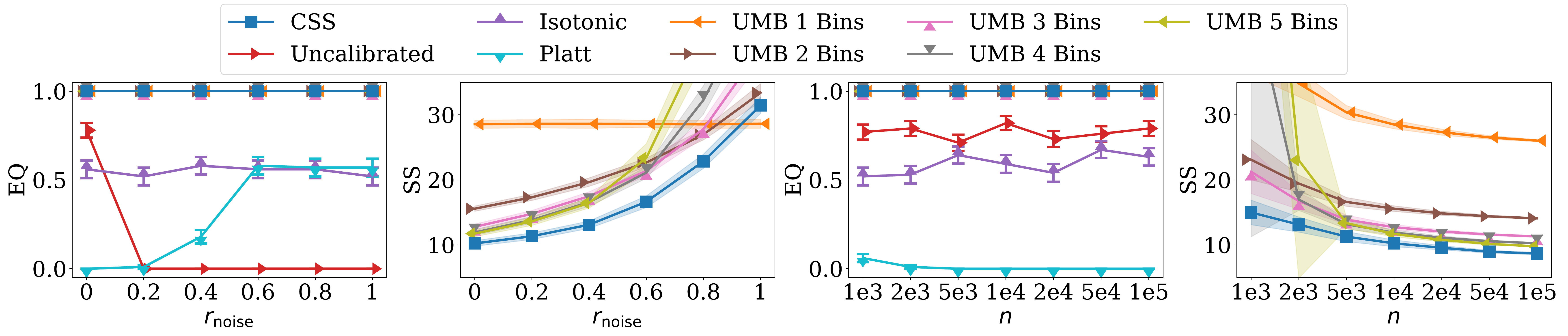}
\caption{Analysis of \algabbr and baselines 
when varying the classifier noise ratio $r_{\text{noise}}$ (\ie, accuracy) and calibration
set sizes $n$.
The first and third plots show the empirical probability that each algorithm provides enough qualified candidates (EQ) 
from $100$ runs with standard error bars (higher is better). 
The second and fourth plots show the average, among $100$ runs with one standard deviation as shaded regions, expected size (SS) 
of the shortlists (lower is better). We do not plot SS for algorithms that fail the quality requirement in terms of EQ.
%
In the left two plots, the size of the calibration set is $n = 10,000$. In the right two plots, the classifier noise ratio
is $r_{\textnormal{noise}} = 0$.
}
\label{fig:exp_normal}
\end{figure*}
Our theoretical results have shown that \algabbr is robust to the accuracy of the classifier and the amount of calibration data.
%
%
However, \algabbr does not account for the potential differences in accuracy or in the amount of calibration data across demographic groups.
As a result, qualified candidates in minority groups may be unfairly underrepresented in the shortlists provided by \algabbrNS.
%

To tackle the above problem and increase the diversity of the shortlists, we design a variant of \algabbrNS, which we name \algdivabbrNS, motivated by the Rooney rule~\cite{collins2007tackling} and multicalibration~\cite{hebert2018multicalibration} (refer to Appendix~\ref{app:screening-diversity}). 
\algdivabbr quantifies the uncertainty
in the estimation of the number of qualified candidates separately for
each demographic group, so that we have distribution-free guarantees for each demographic group. This means that \algdivabbr will include more candidates in the shortlist for groups with higher uncertainty to ensure a sufficient number of qualified candidates from each group.
Effectively, \algdivabbr shifts the cost of uncertainty from minority candidates to the decision maker, since it creates a potentially longer shortlist.
Interestingly, this provides an economic incentive for the decision maker to both build classifiers that perform well across demographic groups and to collect more calibration data about minority groups. 

\vspace{-1mm}
\section{Empirical Evaluation Using Survey Data}
\label{sec:real}
\vspace{-1mm}
%
%

In this section, we compare \algabbr against several competitive baselines on multiple 
instan\-ces of a simulated screening process created using US Census survey data. 
%
%
%
%

\xhdr{Experiment Setup}
We create a simulated screening process using a dataset comprised of employment 
information for $\sim$$3.2$ million individuals from the US Census~\cite{ding2021retiring}. 
%
%
For each individual, we have sixteen features $x \in \RR^{16}$ (\eg, education, marital 
status) and a label $y\in\{0, 1\}$ that indicates whether the individual is
employed ($y = 1$) or unemployed ($y=0$). 
Race is among the features, which we use as the protected attribute as suggested by Ding et al.~\cite{ding2021retiring}. We treat white as the majority group $g_{\textnormal{maj}}$ and all other 
races as the mi\-no\-ri\-ty group $g_{\textnormal{min}}$. 
To ensure that there is a limited number of qualified candidates ($\sim$$20$\%) 
in each of the simulated screening processes, we randomly downsample the 
dataset\footnote{For the diversity experiments, we downsample the majority 
and minority groups independently.}.
%
After downsampling, the dataset contains $\sim$$2.2$ million individuals. 

For the experiments, we randomly split the dataset into two equally-sized and disjoint subsets. 
%
%
We use one of the subsets to create a training set of $10{,}000$ individuals, as well as calibration sets with varying sizes $n$. We use the other subset to simulate pools of candidates for testing. 
%
%
To get a screening classifier, we train a logistic regression $f_{\textnormal{LR}}$ on the training set to 
predict the probability $f_{\textnormal{LR}}(x)$ that a candidate is qualified. 
Here, we 
vary the accuracy of the classifier by replacing, with probability $r_{\textnormal{noise}}$, each of its predictions with some noise $\beta\sim\textnormal{Beta}(1, 4)$, \ie, $f=\gamma\beta + (1-\gamma)f_{\textnormal{LR}}$, where $\gamma\sim\textnormal{Bernoulli}(r_{\textnormal{noise}})$ and $r_{\textnormal{noise}}$ is the \emph{classifier~noise~ratio}. 

In each simulated screening process, we set the size of the test pool of candidates to $m = 100$, 
the desired expected number of qualified candidates to $k = 5$,
and the success probability to $1 - \alpha = 0.9$.
For the diversity experiments, 
%
%
we set the desired expected number of qualified candidates $k_{\textnormal{maj}}$ and $k_{\textnormal{min}}$ so that the equal opportunity constraint,\ie, $k_{\textnormal{maj}} / (m\EE_{(X,Y)}\sbr{Y\II\cbr{X\in g_{\textnormal{maj}}}}) = k_{\textnormal{min}} / (m\EE_{(X,Y}\sbr{Y\II\cbr{X\in g_{\textnormal{min}}}})$
is satisfied~\cite{hardt2016equality} subject to $k_{\textnormal{maj}} +k_{\textnormal{min}} = 5$.
%
%
%
\begin{figure*}[t]
\centering
\includegraphics[width=1.\textwidth]{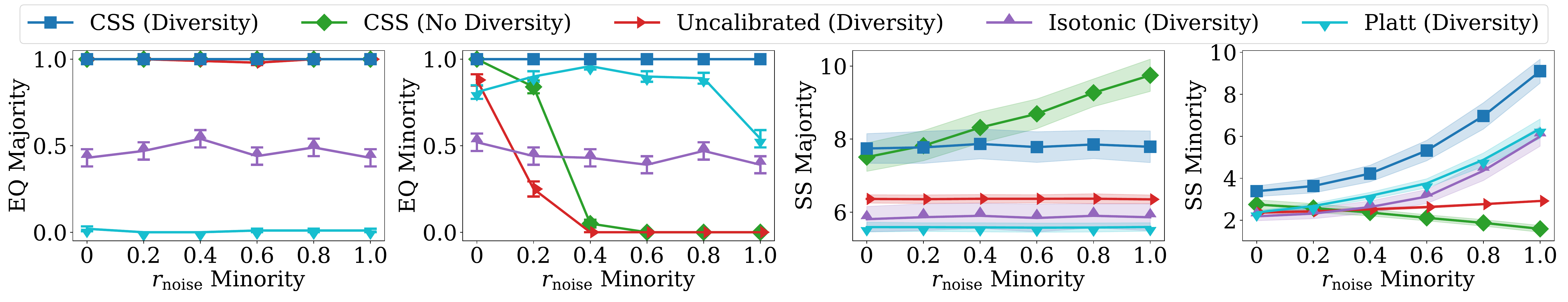}
\caption{Analysis of diversity guarantees of \algdivabbr and several baselines when varying the classifier noise ratio $r_{\text{noise}}$ (\ie, accuracy) for individuals in the minority group.
For individuals in the majority group, the classifier noise ratio is $r_{\text{noise}} = 0$.
The first and third plots focus on the majority group and the second and fourth plots focus on the minority group.
}
\label{fig:exp_diversity}
\end{figure*}

\xhdr{Methods} In our experiments, we compare \algabbr with several 
baselines. 
Since no prior screening algorithms with distribution-free guarantees exist, we introduce 
a simple screening algorithm based on uniform mass binning~\cite{zadrozny2001obtaining} (UMB $B$ Bins), which also enjoys distribution-free guarantees on the expected number of qualified candidates. 
The algorithm bounds the calibration error of a classifier that is calibrated on $B$ bins of 
roughly equal size and selects the candidates from top-scored bins to low-scored bins (and possibly at random from the last bin it selects candidates from) until a lower bound on the expected 
number of qualified candidates is no smaller than $k$. 
Refer to Appendix~\ref{app:screening-multiple-bins} for more details about this baseline algorithm.

We also compare \algabbr with three other baselines that do not provide distribution-free guarantees.
The first is called {\em Uncalibrated}, and it applies the optimal decision rule for omniscient classifiers as if $f$ were the omniscient classifier defined in Eq.~\ref{eq:decision_rule_omniscient_classifier}.
The second is called {\em Platt}, since it first calibrates $f$ using Platt 
scaling~\cite{platt1999probabilistic} and then proceeds like Uncalibrated with the calibrated classifier.
The third is called {\em Isotonic}. It treats the classifier $f$ as a 
regression model from $x$ to $y$, and then employs Isotonic regression calibration~\cite{kuleshov2018accurate} to produce a calibrated regression model $h$ that estimates $h(t)\approx\EE\sbr{(Y\II\cbr{f(x)\geq t}}$. It then selects the largest threshold $t$ such that $mh(t)\geq k$ for shortlisting.

\xhdr{Metrics} To compare the screening algorithms, we run the experiments $100$ times for each algorithm and setting. For each run, we estimate whether each algorithm provides shortlists that contain a large enough expected number of qualified candidates $\textnormal{EQ} = \II\cbr{\hat{\EE}\sbr{\sum_{i\in[m]}Y_iS_i} \geq k}$, and its expected shortlist size $\textnormal{SS} = \hat{\EE}\sbr{\sum_{i\in[m]}S_i}$, using $1,000$ independent pools of candidates sampled at random from the test set. We then compare the algorithms in terms of the percentage of times, along with standard errors, they provide enough qualified candidates. We also compare the average shortlist size, along with standard deviations. 

\vspace{1mm} \noindent {\bf How does the accuracy of the classifier affect different screening algorithms?}
%
%
The left two plots in Figure~\ref{fig:exp_normal} show
how \algabbr compares to the baselines for classifiers of
varying accuracy (\ie, varying $r_{\textnormal{noise}}$). 
The results show that the shortlists provided by the algorithms with distribution-free guarantees (i.e., \algabbr and UMB) do contain enough qualified candidates, while the others fail. 
%
%
Moreover, \algabbr and UMB are robust to the accuracy of the 
classifier they use, as they compensate for decreased accuracy with longer shortlists to maintain their guarantees.
%
In addition, we find that the shortlists provided by these algorithms contain enough qualified candidates more frequently than suggested by the worst-case theoretical guarantee of $1-\alpha=0.9$. Compared to all UMB variants, \algabbr provides  smaller shortlists except for the pure-noise case.
%
%
Note that this exception does not violate the optimality of \algabbr among deterministic threshold policies using the empirical lower bound as shown in Theorem~\ref{eq:decision-threshold-rule-algorithm}, since UMB algorithms are allowed to randomly select candidates in the last bin they select candidates from, as discussed
previously and in Appendix~\ref{app:screening-multiple-bins} in detail. 
%

\vspace{1mm} \noindent {\bf What is the effect of different amounts of calibration data on the screening algorithms?}
%
%
The right two plots in Figure~\ref{fig:exp_normal} show how the screening algorithms perform for increasing amounts of calibration data $n$.
We see that \algabbr and UMB are robust to the amount of calibration data, and can effectively account for less data by increasing the shortlist size to maintain their guarantees. In terms of shortlist size, \algabbr is more effective over the whole range of $n$, since it provides smaller shortlists than all of the UMB variants.

%

\vspace{1mm} \noindent {\bf How does the accuracy of the classifier affect different groups of candidates?}
Figure~\ref{fig:exp_diversity} shows how \algdivabbr and the baselines perform on the majority and minority
groups as we decrease the accuracy of the classifier for individuals in the minority group 
(\ie, we increase $r_{\textnormal{noise}}$ only for individuals in the minority group). 
%
%
Here, \algabbr (No Diversity) refers to naively applying \algabbr
on the entire pool of candidates, without diversity requirements. 
We allow all other algorithms to select candidates from the majority group and the minority 
group separately. 
The results show that, as the accuracy of the classifier for individuals in the minority group decreases, the shortlists provided by \algabbr (No Diversity) contain more and more (fewer and fewer) candidates from the majority (minority) group. 
This suggests that we should explicitly account for diversity in the screening process, especially when the accuracy of the classifier differs across groups. 
While Uncalibrated, Platt and Isotonic select candidates across groups separately, the
shortlists they provide contain fewer qualified candidates from the minority group as 
the accuracy worsens (Uncalibrated) or do not contain enough qualified candidates from 
both groups (Platt, Isotonic).
In contrast, \algdivabbr adapts to the loss in accuracy of the classifier for individuals 
in the minority group and the shortlists it provides contain enough qualified candidates 
from both groups.
We found similar results also when we varied the number of individuals from each group in 
the calibration data, instead of the classifier accuracy.

\vspace{-1mm}
\section{Conclusions}
\label{sec:conclusions}
\vspace{-1mm}
In this work, we initiated the development of screening algorithms that provide distribution-free guarantees on the quality of the shortlisted candidates. 
In particular, we proposed the \algabbr algorithm, which can be applied to any screening classifier. We show that for any amount of calibration data, \algabbr selects near-optimal shortlists of candidates, in terms of the expected shortlist size, while ensuring that the expected number of qualified candidates is above a desired target with high probability. Moreover, we showed how the \algdivabbr variant, which shortlists different groups of candidates separately, can ensure diversity of the shortlists with similar guarantees. Both the theoretical analysis and the empirical evaluation confirm that \algabbr and \algdivabbr are robust to the accuracy of the given classifier and the amount of calibration data. 

Our work opens up many interesting avenues for future work. 
For example,
we have assumed that candidates do not present themselves strategically to the screening algorithms
and that the data distribution at calibration and test time is the same.
It would thus be interesting to develop screening algorithms that are robust to strategic behaviors~\cite{tsirtsis2020decisions} and distribution shifts~\cite{podkopaev2021distribution}.
Moreover, it is crucial to fully understand how screening algorithms can most effectively and most fairly augment (biased) human decision making in real selection processes.

\vspace{-1mm}
\section*{Acknowledgements}
\vspace{-1mm}
We thank Eleni Straitouri and Nastaran Okati for discussions and feedback in the initial stage of this work. 
Gomez-Rodriguez acknowledges the support from the European Research Council (ERC) under the European Union’s Horizon 2020 research and
innovation programme (grant agreement No. 945719). 
Wang and Joachims acknowledge the support from NSF Awards IIS-1901168 and IIS-2008139. 
All content represents the opinion of the authors, which is not necessarily shared or endorsed by their respective employers and/or sponsors. 

{ 
\bibliographystyle{unsrt}
\bibliography{refs}
}

\onecolumn

\appendix

\section{Additional Algorithms}\label{app:algorithms}
\subsection{Calibrated Screening Algorithm under the Dynamic Pool Size Setting}
\label{app:dynamic-pool-size}

\begin{algorithm}[t]
\begin{algorithmic}[1]
\STATE{{\bf input:} $k$, $\EE\sbr{M}$, $\Dcal_{\textnormal{cal}}$, $f$, $\alpha$, $\xb$}
\STATE{{\bf initialize:} $\sbb=\emptyset$ }
\STATE{$\hat{t}_f=\sup\cbr{t\in[0,1]\lgiven \hat{\delta}_{t,1} \geq k/\EE\sbr{M} + \epsilon(\alpha,n)}$}
\FOR{ $x \in \xb$ }
\IF{$f(x)\geq \hat{t}_f$}
\STATE add $x$ to $\sbb$
\ENDIF
\ENDFOR
\STATE{{\bf return} $\sbb$}
\end{algorithmic}
\caption{\algabbr (Dynamic Pool Size)}
\label{alg:screening-set-and-dynamic-size}
\end{algorithm}
In reality, the size of the pool of candidates might be different across times. Here, we derive an algorithm \algabbr (Dynamic Pool Size) illustrated in Algorithm~\ref{alg:screening-set-and-dynamic-size} that allows for dynamic size of the pool of applicants.  we overload the use of notation $\sbb$ to denote the set of the selected candidates rather than a vector, for ease of presentation, especially for the screening algorithm to ensure diversity as we will discuss in the next subsection. \algabbr (Dynamic Pool Size) requires that the expected size $\EE\sbr{M}$ is given, which can be estimated from past pools. We show that all our theoretical results naturally generalize to the dynamic pool size setting. 

In the dynamic pool size setting, instead of assuming that the size of the pool of candidates $m$ is constant, we assume the size is a random variable $M\sim P_{M}$ that follows some distribution $P_{M}$, with mean $\EE\sbr{M}=\EE_{M\sim P_{M}}[M]$. Thus, the data generation process for the pool of candidates becomes $M,\Xb, \Yb\sim P_{M} \times P_{X,Y}^M$. 

For the individual guarantees, Theorem~\ref{thm:omniscient-classifier-individual} naturally holds for all $m\in\RR^+$, and thus for the dynamic size setting; the impossibility results in Proposition~\ref{prop:impossibility-individual-guarantees} still hold in the dynamic pool size setting, since the constant size setting is a special case of the dynamic pool size setting. 

So we focus on showing that the theoretical results in the marginal guarantees still hold here, by showing that there is a family of policies that are oblivious to the size of the pools while representative enough. We call this family of policies pool-independent policies. 
\begin{definition}[Pool-Independent Policies]
Given a classifier $f$, a policy $\pi_f\in\Pi_f$ is pool-independent if there is a mapping $p_{\pi_f} : \range(f) \rightarrow [0,1]$, such that for any candidate $x_i$ in any pool $\xb$ of any size $m$, the probability that the candidate is selected under $\pi_f$ is $\Pr\rbr{S_i=1} = p_{\pi_f}(f(x_i))$. 
\end{definition}
This family of policies have the nice property that their expected number of qualified candidates in the shortlist and the expected shortlist size depend on $P_{M}$ only through $\EE\sbr{M}$. Formally speaking, for a pool-independent policy $\pi_f$ and its selection probability mapping $p_{\pi_f}$,
\[
\EE_{(M, \Xb, \Yb)\sim P, \Sbb\sim \pi_f}\sbr{\sum_{i\in[M]}S_i Y_i} = \EE\sbr{M}\EE_{(f(X), Y) \sim P_{f(X),Y}}\sbr{p_{\pi_f}(f(X))Y}
\]
and 
\[
\EE_{(M, \Xb)\sim P, \Sbb\sim \pi_f}\sbr{\sum_{i\in[M]}S_i} = \EE\sbr{M}\EE_{f(X) \sim P_{f(X)}}\sbr{p_{\pi_f}(f(X))}. 
\]
We show in the following proposition that the family of pool-independent policies are representative enough in the dynamic pool size setting, in a sense that for any given classifier $f$ and any policy $\pi_f\in\Pi_f$, we can always find a pool-independent policy in $\Pi_f$ that achieves the same expected number of qualified candidates in the shortlists and the same expected size of the shortlists as $\pi_f$. 
\begin{proposition}
\label{prop:pool-independent-policy-representative}
Given a classifier $f$, for any policy $\pi_f\in\Pi_f$, there exists a pool-independent policy $\pi_f'\in\Pi_f$ such that they have the same expected number of qualified candidates
\[
\EE_{(M,\Xb,\Yb)\sim P, \Sbb\sim\pi}\sbr{\sum_{i\in[M]}Y_iS_i} = \EE_{(M,\Xb,\Yb)\sim P, \Sbb\sim\pi'}\sbr{\sum_{i\in[M]}Y_iS_i} 
\]
and the same expected shortlist size 
\[
\EE_{(M,\Xb)\sim P, \Sbb\sim\pi}\sbr{\sum_{i\in[M]}S_i} = \EE_{(M,\Xb)\sim P, \Sbb\sim\pi'}\sbr{\sum_{i\in[M]}S_i}. 
\]
\end{proposition}

This implies that it is sufficient to find optimal/near-optimal solutions within this family of pool-independent policies to achieve global optimal/near-optimal guarantees. In fact, all the optimal/near-optimal policies in all the theorems, propositions, corollaries are all pool-independent policies. They all can be translated to the dynamic pool size setting by setting $m=\EE\sbr{M}$ in the policy design. 

\subsection{Calibrated Subset Selection Algorithm to Ensure Diversity}
\label{app:screening-diversity}

\begin{algorithm}[t]
\begin{algorithmic}[1]
\STATE{{\bf input:} $\Gcal$, $\cbr{\EE\sbr{M}_g}_{g\in\Gcal}$, $\cbr{k_g}_{g\in\Gcal}$, $\cbr{\Dcal^g_{\textnormal{cal}}}_{g\in\Gcal}$, $f$, $\alpha$, $\cbr{\xb_g}_{g\in\Gcal}$}
\FOR{ $g\in\Gcal$ }
\STATE $\sbb_g=$ \algabbr (Dynamic Pool Size) ($k_g$, $\EE\sbr{M}_g$, $\Dcal^g_{\textnormal{cal}}$, $f$, $\alpha/\left|\Gcal\right|$, $\xb_g$)
\ENDFOR
\STATE{{\bf return} $\cup_{g\in\Gcal}\sbb_g$}
\end{algorithmic}
\caption{\algdivabbr}
\label{alg:screening-diversity}
\end{algorithm}

\algdivabbr (Algorithm~\ref{alg:screening-diversity}) selects calibrated shortlists of candidates for each demographic group independently. \algdivabbr requires inputs of the groups $\Gcal$, the expected pool size of each group $\cbr{\EE\sbr{M}_g}_{g\in\Gcal}$, the target expected number of qualified candidates for each group $\cbr{k_g}_{g\in\Gcal}$, the calibration data for each group $\cbr{\Dcal^g_{\textnormal{cal}}}_{g\in\Gcal}$, and the pool of candidates from each group $\cbr{\xb_g}_{g\in\Gcal}$. Note that the groups can have overlap, \ie, each candidate might belong to multiple groups. The guarantees on the expected number of qualified candidates from \algabbr (Dynamic Pool Size) directly apply to each group. The near-optimality guarantees of \algabbr (Dynamic Pool Size) only apply for every group when each candidate belongs to exactly one group, since otherwise it is possible that we select more-than-enough candidates from a particular group to satisfy the guarantees for the other groups. 

\subsection{Calibrated Screening Algorithms with Multiple Bins}
\label{app:screening-multiple-bins}

\begin{algorithm}[t]
\begin{algorithmic}[1]
\STATE{{\bf input:} $k$, $m$, $B$, $\Bcal$, $\cbr{\hat{\delta}_{b}}_{b\in[B]}$, $\epsilon(\alpha, n)$, $\xb$}
\STATE{{\bf initialize:} $\sbb=\emptyset$ }
\STATE{$\hat{b}=\inf\cbr{a\in[B]\lgiven \sum_{b\in[a]}\rbr{\hat{\delta}_{b} - 2\epsilon(\alpha, n)} \geq k/m}$}
\STATE{$\hat{\theta} = \frac{k/m - \sum_{b\in[\hat{b}-1]  } \rbr{\hat{\delta}_b - 2\epsilon(\alpha, n)}}{\hat{\delta}_{\hat{b}} - 2\epsilon(\alpha, n)}$}
\FOR{ $x \in \xb$ }
\IF{$\Bcal(x) < \hat{b}$}
\STATE add $x$ to $\sbb$
\ELSIF{$\Bcal(x) = \hat{b}$}
\STATE add $x$ to $\sbb$ with probability $\hat{\theta}$
\ENDIF
\ENDFOR
\STATE{{\bf return} $\sbb$}
\end{algorithmic}
\caption{\algname with Multiple Bins}
\label{alg:screening-multiple-bins}
\end{algorithm}

We develop algorithms that can ensure distribution-free guarantees on the expected number of qualified candidates in the shortlists, when we partition the sample-space into multiple bins by the prediction scores from a given classifier $f$. For ease of notation and presentation of the proof, we illustrate the theory and the algorithm in the constant pool size setting. But they can be similarly adapted to the dynamic pool size setting as described in Appendix~\ref{app:dynamic-pool-size}. 

Let $t_0=1,\ldots,t_{B} = 0$ be the thresholds on the prediction scores from $f$ in decreasing order, which partition the sample-space $\Xcal$ into $B$ bins: $\cbr{x\in\Xcal \given t_0 \leq f(x) \leq t_1}$, $\cbr{x\in\Xcal \given t_1 < f(x) \leq t_2}$, \ldots, $\cbr{x\in\Xcal \given t_{B-1} < f(x) \leq t_{B}}$. Let $\Bcal:\Xcal\rightarrow[B]$ be the bin identity function, which maps a candidate $x$ to the bin it belongs to $\Bcal(x)$. Let $\delta_{b}=\EE_{(X,Y)\sim P_{X,Y}}\sbr{\II\cbr{\Bcal(X) = b}Y}$, and $\hat{\delta}_b = \frac{1}{n}\sum_{i\in[n]}Y_i^c\II\cbr{\Bcal(X_i^c)}$ be an empirical estimate of it from the calibration data $\Dcal_{\textnormal{cal}}$. There exist many concentration inequalities to bound the calibration errors. But they generally require sample-splitting, \ie, to split the calibration data into two sets, one set for selecting the sample-space partition, and the other set for estimating the calibration error. A recent work~\cite{gupta2021distribution} on distribution-free calibration without sample-splitting does not apply here, since their proposed method can only bound the calibration errors of $\mu_b=\EE_{(X,Y)\sim P_{X,Y}}\sbr{Y_i \given \Bcal(X) = b}$, rather than $\delta_b$. To avoid sample-splitting while still bounding the calibration errors of $\delta_b$ on sample-space partitions with thresholds $\cbr{t_{b}}_{b\in[B-1]}$ chosen in a data dependent way (\eg, uniform mass binning~\cite{zadrozny2001obtaining}), we again use the DKWM inequality to bound the calibration error of $\delta_b$ on any bin $b$ that corresponds to an interval on the scores predict by $f$, by the following proposition. 
\begin{proposition}
\label{prop:calibration-errors-multiple-bins}
For any $\alpha\in(0,1)$, with probability at least $1-\alpha$ (in $f$ and $\Dcal_{\textnormal{cal}}$), it holds that for any bin $\cbr{x\in\Xcal\given t_{b-1}\leq f(x)\leq t_b}$ indexed by $b$ and parameterized by $0\leq t_{b-1} < t_{b}\leq 1$, 
\[
\left|\delta_b - \hat{\delta}_b\right| \leq \sqrt{2\ln(2/\alpha)/n} = 2\epsilon(\alpha,n). 
\]
\end{proposition}

With this calibration error guarantee that holds for any bin, and motivated by the monotone property of $f$, we derive an empirical lower bound on the expected number of qualified candidates similarly as Corollary~\ref{cor:bounds_on_objectives} using a family of threshold decision policies that select candidates from the top-scored bins to lower-scored bins, possibly at random for the candidates in the last bin it selects candidates from. Any policy $\pi_{f,a,\theta}$ given a classifier $f$ in this family makes decisions based on the following decision rule
\begin{equation*}
s_i =
\begin{cases}
1 & \textnormal{if} \, \Bcal(x_i) < a, \\
\textnormal{Bernoulli}(\theta) & \textnormal{if} \, \Bcal(x_i) = a, \\
0 & \textnormal{otherwise}.
\end{cases}
\end{equation*}

An empirical lower bound on the expected number of qualified candidates in the shortlists can be derived as follows
\begin{align*}
\EE_{(\Xb, \Yb) \sim P, \Sbb \sim \pi_{f,a, \theta}} \sbr{\sum_{i\in[m]}S_i Y_i} &= m\EE_{(\Bcal{(X), Y)}\sim P_{\Bcal(X), Y}}\sbr{Y\rbr{\II\cbr{\Bcal(X) < a} + \theta \II\cbr{\Bcal(X) = a}}}\\
&=m\sum_{b\in[B]}\delta_b\rbr{\II\cbr{b < a} + \theta\II\cbr{b = a}}\\
&\geq m\sum_{b\in[B]}\rbr{\hat{\delta}_b - 2\epsilon(\alpha, n)}\rbr{\II\cbr{b < a} + \theta\II\cbr{b = a}}. 
\end{align*}

We solve the optimization problem with this empirical lower bound similarly as in Eq.~\ref{eq:calibration_from_data_threshold}:
\begin{equation} \label{eq:optimization_problem_calibration_from_data_multiple_bins}
\hat{b}, \hat{\theta} = \argmin_{b, \theta \in \Theta} \, \EE_{\Xb\sim P, \Sbb\sim \pi_{f,b, \theta}}\left[\sum_{i\in[m]}S_i\right]
\end{equation}
where $\Theta = \{ a\in[B], \theta \in [0,1) \given m\sum_{b\in[B]}\rbr{\hat{\delta}_b - 2\epsilon(\alpha, n)}\rbr{\II\cbr{b < a} + \theta\II\cbr{b = a}}\geq k \}$. 

And it turns out that there is a closed form solution
\[
\hat{b}=\inf\cbr{a\in[B]\lgiven \sum_{b\in[a]}\rbr{\hat{\delta}_{b} - 2\epsilon(\alpha, n)} \geq k/m}
\]
and 
\[
\hat{\theta} = \frac{k/m - \sum_{b\in[\hat{b}-1]  } \rbr{\hat{\delta}_b - 2\epsilon(\alpha, n)}}{\hat{\delta}_{\hat{b}} - 2\epsilon(\alpha, n)}. 
\]
We summarize the algorithm using this decision policy in Algorithm~\ref{alg:screening-multiple-bins}. It guarantees that the expected number of qualified candidates is greater than $k$ with probability $1-\alpha$ as shown in the empirical lower bound. And, it also satisfies the near-optimality guarantees on the expected size of the shortlist among this extended family of threshold decision policies given a fixed sample-space partition similarly as \algname does in Proposition~\ref{prop:gap_calibration_from_data}. We omit the proof since it can be simply adapted from the proof of Proposition~\ref{prop:gap_calibration_from_data}. 

The disadvantage of Algorithm~\ref{alg:screening-multiple-bins} is that it requires a given sample-space partition and finds near-optimal policy only among the policies using this fixed sample-space partition, while \algabbr directly optimizes the sample-space partition and the policy jointly. To also optimize over sample-space partitions, \ie, the number of bins and the thresholds, is computationally intractable and hard to identify which one is actually optimal. \algabbr solves this problem by considering a smaller policy space (no random selection from the last bin) and the sample-space partition space (only two bins), so that it is easy to find the optimal sample-space partition and the optimal policy simultaneously. We now show that among deterministic threshold decision rules, sample-space partitions with two bins are optimal. 

\xhdr{Calibration on More Bins Worsens the Performance of Threshold Policies} 
When we consider deterministic threshold decision rules, Algorithm~\ref{alg:screening-multiple-bins} becomes selecting the candidates in bin $\hat{b}$ with probability $1$ instead of with probability $\hat{\theta}$, to ensure distribution-free guarantees on the expected number of qualified candidates in the shortlists. This is equivalent to finding the largest threshold $\hat{t}$ among $\cbr{t_{b}}_{b\in[B]}$ to ensure enough qualified candidates
\[
\hat{t} = \sup\cbr{t\in \cbr{t_{b}}_{b\in[B]} \lgiven \sum_{b\in[B]}\II\cbr{t_b \geq t}(\hat{\delta_b} - 2\epsilon(\delta, n)) \geq k}. 
\]

This is essentially solving the optimization problem using the empirical lower bound on the expected number of qualified candidates
\[
\sum_{b\in[B]}\II\cbr{t_b \geq t}(\hat{\delta_b} - 2\epsilon(\delta, n)), 
\]
over a subset of thresholds in $[0,1]$. This empirical lower bound is looser compared to the bound used in \algabbrNS, which also optimizes over the entire threshold space $[0,1]$, as shown in the following
\[
\sum_{b\in[B]}\II\cbr{t_b \geq t}(\hat{\delta_b} - 2\epsilon(\delta, n)) =  \delta_{t_b, 1} -2\sum_{b\in[B]}\II\cbr{t_b \geq t} \epsilon(\alpha,n) > \delta_{t_b,1} - \epsilon(\alpha,n). 
\]
So it is clear that \algabbr which uses a tighter bound over a larger space of thresholds achieves shorter shortlists in expectation. 


\xhdr{UMB $B$ Bins Algorithms} In the empirical evaluation, the UBM $B$ Bins algorithms correspond to selecting the thresholds such that the calibration data $\Dcal_{\textnormal{cal}}$ are partitioned evenly into $B$ bins, and running Algorithm~\ref{alg:screening-multiple-bins} with the sample-space partition characterized by these thresholds.

\section{Calibration Error Bounds Imply Distribution-Free Average Calibration in Regression}
\label{app:error-bounds-regression-calibration}
If we regard the given classifier $f$ as a regression model from $x$ to $y$, then the calibration error bounds in Proposition~\ref{prop:bounds_errors_calibration} directly imply a classifier that (approximately) achieves average calibration for any interval~\cite{gneiting2007probabilistic,kuleshov2018accurate}. Translating average calibration in regression to our setting, where the range of the regression model has only two values $\{0, 1\}$, a classifier $h$ is average-calibrated if for any $0\leq t\leq 1$ 
\[
\EE_{(X,Y)\sim P_{X,Y}}\sbr{Y\II\cbr{ h(X) \geq t}} = 1 - t.
\]

From the calibration error guarantees in Proposition~\ref{prop:bounds_errors_calibration}, we know that for any $\delta\in(0,1)$, with probability $1-\delta$, for any $t\in[0,1]$,
\[
\left|\EE_{(X,Y)\sim P_{X,Y}}\sbr{Y\II\cbr{f(X)\geq t}} - \hat{\delta}_{t,1}\right|\leq \epsilon(\alpha,n). 
\]
Let $g(t)=\hat{\delta}_{t,1}$, which is a monotonically decreasing function. Let $t = g^{-1}(1-t')$ in the above inequalities, we get for any $t'\in [0,1]$, 
\[
\left|\EE_{(X,Y)\sim P_{X,Y}}\sbr{Y\II\cbr{f(X)\geq g^{-1}(1-t')}} - (1-t')\right|\leq \epsilon(\alpha,n). 
\]
Since g is monotone, we can get that for any $t\in[0,1]$,
\[
\left|\EE_{(X,Y)\sim P_{X,Y}}\sbr{Y\II\cbr{(g\circ f)(X)\geq 1-t}} - (1-t)\right|\leq \epsilon(\alpha,n). 
\]
Let $\hat{h} = g\circ f$, $\hat{h}$ is approximately calibrated in a sense that with probability $1-\alpha$, for any $t\in[0,1]$, 
\[
1-t-\epsilon(\alpha,\delta)\leq\EE_{(X,Y)\sim P_{X,Y}}\sbr{Y\II\cbr{\hat{h}(X)\geq t}} \leq 1-t+\epsilon(\alpha,\delta). 
\]

\section{Proofs} \label{app:proofs}
\subsection{Proof of Theorem~\ref{thm:omniscient-classifier-individual}}
\label{app:thm-omniscient-classifier-individual}
We first show that for any $\xb\in\Xcal^m$, the constraint in the minimization problem in Eq.~\ref{eq:oracle-policy-individual} is satisfied with equality under $\pi^\star_{f^\star}$, 
\begin{align*}
\EE_{\Sbb\sim\pi^\star_{f^\star},\Yb\sim P}\sbr{\sum_{i\in[m]}S_iY_i}
&=\EE_{\{Y_i\}_{i\in[m]}\sim\prod_{i\in[m]}P_{Y\mid X=x_i},\Sbb\sim \pi\left(\{f(x_i)\}_{i\in[m]}\right)}\left[\sum_{i\in[m]}S_i Y_i\right]\\
&=\sum_{i\in[m]}\Pr\rbr{Y=1\mid X=x_i}\left( \II\left\{f(x_i)>t^\star\right\} + \II\left\{f(x_i) = t^\star\right\}\theta^\star\right)\\
&=\sum_{i\in[m]}f(x_i)\left( \II\left\{f(x_i)>t^\star\right\} + \II\left\{f(x_i) = t^\star\right\}\theta^\star\right)\\
&= k. 
\end{align*}
The last equality is by the definition of $t^\star$ and $\theta^\star$. Next, we will show that it is an optimal solution for any $\xb\in\Xcal^m$ by contradiction. 

The objective in Eq. ~\ref{eq:oracle-policy-individual} using $\pi^\star_{f^\star}$ is 
\begin{align*}
\EE_{\Sbb\sim\pi^\star_{f^\star}} \left[\sum_{i\in[m]}S_i\right]
= \sum_{i\in[m]}\II\left\{\Pr\rbr{Y=1\mid X=x_i}>t^\star\right\} + \theta^\star\II\left\{\Pr\rbr{Y=1\mid X=x_i}=t^\star\right\}. 
\end{align*}

Suppose a policy $\pi'$ achieves smaller objective than $\pi^\star_{f^\star}$ for some $\xb\in\Xcal^m$, \ie,
\begin{align*}
\EE_{\Sbb\sim\pi'\rbr{\cbr{f(x_i)}_{i\in[m]}}}\left[\sum_{i\in[m]}S_i\right]\coloneqq R_{\pi'}(\xb)
< \sum_{i\in[m]}\II\left\{\Pr\rbr{Y=1\mid X=x_i}>t^\star\right\} + \theta^\star\II\left\{\Pr\rbr{Y=1\mid X=x_i}=t^\star\right\}. 
\end{align*}
We will show that the constraint in the optimization problem in Eq.~\ref{eq:oracle-policy-individual} for $\xb$ is not satisfied using $\pi'$. Let
\[
t' \coloneqq \sup\left\{t\in[0,1]:\sum_{i\in[m]}\II\left\{\Pr\rbr{Y=1\mid X=x_i}\geq t\right\}\geq R_{\pi'}(\xb)\right\}
\]
and
\[
\theta' \coloneqq \frac{R_{\pi'}(\xb)-\sum_{i\in[m]}\II\left\{\Pr\rbr{Y=1\mid X=x_i}>t'\right\}}{\sum_{i\in[m]}\II\left\{\Pr\rbr{Y=1\mid X=x_i}=t'\right\}}. 
\]
Note that $t^\star\leq t'$ since
\begin{multline*}
\sum_{i\in[m]}\II\left\{\Pr\rbr{Y=1\mid X=x_i}\geq t^\star\right\}\\
\geq \sum_{i\in[m]}\II\left\{\Pr\rbr{Y=1\mid X=x_i}>t^\star\right\} + \theta^\star\II\left\{\Pr\rbr{Y=1\mid X=x_i}=t^\star\right\}>R_{\pi'}(\xb). 
\end{multline*}
Now, we can show that the constraint is not satisfied using $\pi'$. The left hand side of the constraint using $\pi'$ can be bounded as 
\begin{align*}
\EE_{\Sbb\sim\pi'}\left[\sum_{i\in[m]}Y_i S_i\right] &=\EE_{\{Y_i\}_{i\in[m]}\sim\prod_{i\in[m]}P_{Y\mid X=x_i},\Sbb\sim \pi'}\left[\sum_{i\in[m]}Y_iS_i\right]\\
&=\EE_{\Sbb\sim\pi'}\left[\sum_{i\in[m]}\Pr(Y=1\mid X=x_i)S_i\right]\\
&\leq\sum_{i\in[m]}\Pr(Y=1\mid X=x_i)\left(\II\left\{\Pr(Y=1\mid X=x_i)>t'
\right\} + \II\left\{\Pr(Y=1\mid X=x_i)=t'
\right\}\theta'\right), 
\end{align*}
where the last inequality is by the definition of $R_{\pi'}(\xb)$, $t'$ and $\theta'$. When $t^\star<t'$, 
\begin{multline*}
\sum_{i\in[m]}\Pr\rbr{Y=1\mid X=x_i}\left(\II\left\{\Pr\rbr{Y=1\mid X=x_i}>t'
\right\} + \II\left\{\Pr\rbr{Y=1\mid X=x_i}=t'
\right\}\theta'\right)\\
<\sum_{i\in[m]}\Pr\rbr{Y=1\mid X=x_i}\left( \II\left\{\Pr\rbr{Y=1\mid X=x_i}>t^\star\right\} + \II\left\{\Pr[Y=1\mid X= x_i] = t^\star\right\}\theta^\star\right)
=k. 
\end{multline*}
When $t^\star=t'$,
\[
\theta' = \frac{R_{\pi'}-\sum_{i\in[m]}\II\left\{\Pr\rbr{Y=1\mid X=x_i}>t^\star\right\}}{\sum_{i\in[m]}\II\left\{\Pr\rbr{Y=1\mid X=x_i}=t^\star\right\}}<\theta^\star. 
\]
Thus
\begin{multline*}
\sum_{i\in[m]}\Pr\rbr{Y=1\mid X=x_i}\left(\II\left\{\Pr\rbr{Y=1\mid X=x_i}>t'
\right\} + \II\left\{\Pr\rbr{Y=1\mid X=x_i}=t'
\right\}\theta'\right)\\
=\sum_{i\in[m]}\Pr\rbr{Y=1\mid X=x_i}\left(\II\left\{\Pr\rbr{Y=1\mid X=x_i}>t^\star
\right\} + \II\left\{\Pr\rbr{Y=1\mid X=x_i}=t^\star
\right\}\theta'\right)\\
<\sum_{i\in[m]}\Pr\rbr{Y=1\mid X=x_i}\left(\II\left\{\Pr\rbr{Y=1\mid X=x_i}>t^\star
\right\} + \II\left\{\Pr\rbr{Y=1\mid X=x_i}=t^\star
\right\}\theta^\star\right) =k. 
\end{multline*}
This shows that for any $\xb\in\Xcal^m$, for any policy $\pi'$ that achieves smaller objective than $\pi^\star_{f^\star}$, $\pi'$ does not satisfy the constraint, which concludes the proof. 

\subsection{Proof of Proposition~\ref{prop:impossibility-individual-guarantees}}
\label{app:prop-impossibility-individual-guarantees}
We construct two pools of candidates such that any policy will fail on at least one them.

The two pools of candidates are all $a$s and all $b$s, $\xb =\{a,a,\ldots,a\}$ and $\xb' = \{b,b,\ldots,b\}$. And also note that in this kind of candidate distribution, the only perfectly calibrated predictor that is not the omniscient predictor is the predictor that predicts $h(x) = \Pr\rbr{Y=1}$ for all $x\in\Xcal$, since there are only two possible candidate feature vectors. For any policy $\pi_h$ that makes decisions $S$ based purely on the quality scores by predicted by $h$, the distribution of $\Sbb$ will be the same for the two pools of candidates, since they have the same predictions from $h$. 

Thus the expectation on the size of the shortlist for the two pools of candidates are the same 
\[
\EE_{\Sbb\sim\pi_h\rbr{\cbr{h\rbr{x_{i}}}_{i\in[m]}}}\left[\sum_{i\in[m]}S_i\right] = \EE_{\Sbb\sim\pi_h\rbr{\cbr{h\rbr{x'_{i}}}_{i\in[m]}}}\left[\sum_{i\in[m]}S_i\right] \coloneqq c_{\pi_h}. 
\]
On the other hand, for the optimal policy $\pi^\star_{f^\star}$, we know that 
\[
\EE_{\Sbb\sim\pi^\star_{f^\star}\rbr{\cbr{f^\star\rbr{x_{i}}}_{i\in[m]}}}\left[\sum_{i\in[m]}S_i\right] = k,
\]
and 
\[
\EE_{\Sbb\sim\pi^\star\rbr{\cbr{f^\star\rbr{x'_{i}}}_{i\in[m]}}}\left[\sum_{i\in[m]}S_i\right]= m. 
\]

The sum of the differences on the two pools of candidates is
\begin{multline*}
\left|\EE_{\Sbb\sim\pi_h\rbr{\cbr{h\rbr{x_{i}}}_{i\in[m]}}}\left[\sum_{i\in[m]}S_i\right]-\EE_{\Sbb\sim\pi^\star_{f^\star}\rbr{\cbr{f^\star\rbr{x_{i}}}_{i\in[m]}}}\left[\sum_{i\in[m]}S_i\right]\right|\\
+ \left|\EE_{\Sbb\sim\pi_h\rbr{\cbr{h\rbr{x'_{i}}}_{i\in[m]}}}\left[\sum_{i\in[m]}S_i\right]-\EE_{\Sbb\sim\pi^\star\rbr{\cbr{f^\star\rbr{x'_{i}}}_{i\in[m]}}}\left[\sum_{i\in[m]}S_i\right]\right|
= \left|c_{\pi_h}-k\right| + \left|c_{\pi_h}-m\right|\geq m-k. 
\end{multline*}
So for any policy $\pi_h$, there exists a pool of candidates (either $\xb$ or $\xb'$) such that the difference is at least $\frac{m-k}{2}$. 

For the expected number of qualified candidates in the shortlists, for any policy $\pi_h$, we have 
\[
\EE_{\Yb\sim P, \Sbb\sim\pi_h\rbr{\cbr{h\rbr{x_{i}}}_{i\in[m]}}}\left[\sum_{i\in[m]}S_iY_i\right] = c_{\pi_h}, 
\]
and 
\[
\EE_{\Yb\sim P, \Sbb\sim\pi_h\rbr{\cbr{h\rbr{x'_{i}}}_{i\in[m]}}}\left[\sum_{i\in[m]}S_iY_i\right] =\frac{k}{m} c_{\pi_h}. 
\]
On the other hand, for the optimal policy $\pi^\star_{f^\star}$, we know that 
\[
\EE_{\Yb\sim P, \Sbb\sim\pi^\star_{f^\star}\rbr{\cbr{f^\star\rbr{x_{i}}}_{i\in[m]}}}\left[\sum_{i\in[m]}S_iY_i\right] = \EE_{\Yb\sim P, \Sbb\sim\pi^\star_{f^\star}\rbr{\cbr{f^\star\rbr{x'_{i}}}_{i\in[m]}}}\left[\sum_{i\in[m]}S_iY_i\right] = k. 
\]
The sum of differences on the two pools of candidates is
\begin{multline*}
\left|\EE_{\Yb\sim P, \Sbb\sim\pi_h\rbr{\cbr{h\rbr{x_{i}}}_{i\in[m]}}}\left[\sum_{i\in[m]}S_iY_i\right]-\EE_{\Yb\sim P, \Sbb\sim\pi^\star_{f^\star}\rbr{\cbr{f^\star\rbr{x_{i}}}_{i\in[m]}}}\left[\sum_{i\in[m]}S_iY_i\right] \right|\\
+ \left|\EE_{\Yb\sim P, \Sbb\sim\pi_h\rbr{\cbr{h\rbr{x'_{i}}}_{i\in[m]}}}\left[\sum_{i\in[m]}S_iY_i\right]-\EE_{\Yb\sim P, \Sbb\sim\pi^\star_{f^\star}\rbr{\cbr{f^\star\rbr{x'_{i}}}_{i\in[m]}}}\left[\sum_{i\in[m]}S_iY_i\right] \right| \\
=\left|c_{\pi_h}-k\right|+ \left|\frac{k}{m}c_{\pi_h}-k\right|
\geq k\left(1-\frac{k}{m}\right),
\end{multline*}
where the equality is achieved when $c_{\pi}=k$. 
So for any policy $\pi_h$, there exist a set of candidates (either $\xb$ or $\xb'$) such that the difference is at least $\frac{k}{2}\left(1-\frac{k}{m}\right)$. 


\subsection{Proof of Theorem~\ref{thm:perfectly-calibrated-classifier-marginal}}
\label{app:thm-perfectly-calibrated-classifier-marginal}
First, we show that the constraint in the minimization problem in~\eqref{eq:oracle-policy-marginal} is satisfied with equality: 
\begin{align*}
    \EE_{(\Xb,\Yb)\sim P, \Sbb\sim\pi^\star_h}\left[\sum_{i\in[m]}Y_iS_i\right]&=\EE_{\cbr{(h(X_i),Y_i)}_{i\in[m]}\sim P_{h(X),Y}^m,\Sbb\sim \pi^\star_h\left(\{h(X_i)\}_{i\in[m]}\right)}\left[\sum_{i\in[m]}Y_i S_i\right]\\
    &=\EE_{\{h(X_i)\}_{i\in[m]}\sim P_{h(X)}^m,\Sbb\sim \pi^\star_h\left( \{h(X_i)\}_{i\in[m]}\right)}\left[\sum_{i\in[m]}S_i\Pr\rbr{Y=1\mid h(X_i)}\right]\\
    &=\EE_{\{h(X_i)\}_{i\in[m]}\sim P_{h(X)}^m,\Sbb\sim \pi^\star_h\left( \{h(X_i)\}_{i\in[m]}\right)}\left[\sum_{i\in[m]}S_ih(X_i)\right]\\
    &=\EE_{\{h(X_i)\}_{i\in[m]}\sim P_{h(X)}^m}\left[\sum_{i\in[m]}\left[\II\left\{h(X_i)>t_h\right\}h(X_i) + \II\left\{h(X_i)=t_h\right\} h(X_i)\theta_h\right] \right]\\
    &=m\EE_{h(X)\sim P_{h(X)}}\left[\II\left\{h(X)>t_h\right\}h(X) + \II\left\{h(X)=t_h\right\} h(X)\theta_h\right]\\
    &= k, 
\end{align*}
where the last equality is by the definition of $t_h$ and $\theta_h$. 

Next, we will show that it is an optimal solution by contradiction. The objective in~\eqref{eq:oracle-policy-marginal} using $\pi^\star_h$ is 
\begin{align*}
\EE_{\Xb\sim P, \Sbb\sim\pi^\star_h}\left[\sum_{i\in[m]}S_i\right]
&= \EE_{\Xb\sim P}\left[\sum_{i\in[m]}\left(\II\left\{h(X_i)>t_h\right\} + \theta_h\II\left\{h(X_i)=t_h\right\}\right)\right]\\
&= \EE_{\{h(X_i)\}_{i\in[m]}\sim P_{h(X)}^m}\left[\sum_{i\in[m]}\left(\II\left\{h(X_i)>t_h\right\} + \theta_h\II\left\{h(X_i)=t_h\right\}\right)\right]\\
&= m\EE_{h(X)\sim P_{h(X)}}\left[\II\left\{h(X)>t_h\right\} + \theta_h\II\left\{h(X)=t_h\right\}\right]\\
&= m\sum_{b\in[B]}\rho^\star_b\left(\II\left\{\mu^\star_b>t_h\right\} + \theta_h\II\left\{\mu^\star_b = t_h\right\}\right). 
\end{align*}

For any policy $\pi'_h$ that achieves smaller objective than $\pi^\star_h$
\begin{align*}
\EE_{\Xb\sim P, \Sbb\sim\pi'_h}\left[\sum_{i\in[m]}S_i\right]\coloneqq R_{\pi'_h}< m\sum_{b\in[B]}\rho^\star_b\left(\II\left\{\mu^\star_b>t_h\right\} + \theta_h\II\left\{\mu^\star_b = t_h\right\}\right), 
\end{align*}
we will show that the constraint is not satisfied using $\pi'_h$. Let
\[
t'_h \coloneqq \sup\left\{t\in[0,1]:\sum_{b\in[B]}\rho^\star_b\II\left\{\mu^\star_b\geq t\right\}\geq R_{\pi'_h}/m\right\}, 
\]
and
\[
\theta' = \frac{R_{\pi'_h}/m-\sum_{b\in[B]}\rho^\star_b\II\left\{\mu^\star_b>t'\right\}}{\sum_{b\in[B]}\rho^\star_b\II\left\{\mu^\star_b=t'\right\}}. 
\]
Note that $t_h\leq t'$ since
\begin{align*}
\sum_{b\in[B]}\rho_b^\star\II\left\{\mu^\star_b\geq t_h\right\}\geq \sum_{b\in[B]}\rho_b^\star\left(\II\left\{\mu^\star_b> t_h\right\} + \theta_h\II\left\{\mu^\star_b= t_h\right\}\right) > R_{\pi'_h}/m. 
\end{align*}
Now we can show that the constraint is not satisfied using $\pi'_h$. The left hand side of the constraint using $\pi'_h$ can be bounded as 
\begin{align*}
    \EE_{(\Xb,\Yb)\sim P, \Sbb\sim \pi'_h}\left[\sum_{i\in[m]}Y_iS_i\right]
    &=\EE_{\{(h(X_i),Y_i)\}_{i\in[m]}\sim P_{h(X),Y}^m,\Sbb\sim \pi'_h\left(\cbr{h\rbr{X_i}}_{i\in[m]}\right)}\left[\sum_{i\in[m]}Y_iS_i\right]\\
    &=\EE_{\cbr{h(X_i)}_{i\in[m]}\sim P^m_{h(X)}, \Sbb\sim\pi'_h\rbr{\cbr{h(X_i)}_{i\in[m]}}}\sbr{\sum_{i\in[m]}S_i\Pr\rbr{Y=1\mid h(X_i)}}\\
    &=\EE_{\cbr{h(X_i)}_{i\in[m]}\sim P^m_{h(X)}, \Sbb\sim\pi'_h\rbr{\cbr{h(X_i)}_{i\in[m]}}}\sbr{\sum_{i\in[m]}S_ih(X_i)}\\
    &\leq m\EE_{h(X)\sim P_{h(X)}}\left[h(X)\left(\II\left\{h(X)>t'_{h}\right\} + \theta'_h\II\left\{h(X)=t'_h\right\}\right)\right]\\
    &= m\sum_{b\in[B]}\rho^\star_b\mu^\star_b\left(\II\left\{\mu^\star_b>t'_h\right\} + \theta'_h\II\left\{\mu^\star_b=t'_h\right\}\right). 
\end{align*}
The inequality is by the definition of $R_{\pi'_h}$, $t'_h$ and $\theta'_h$. When $t_h<t'_h$, 
\[
m\sum_{b\in[B]}\rho^\star_b\mu^\star_b\left(\II\left\{\mu^\star_b>t'_h\right\} + \theta'_h\II\left\{\mu^\star_b=t'_h\right\}\right)
m\sum_{b\in[B]}\rho^\star_b\mu^\star_b\left(\II\left\{\mu^\star_b>t_h\right\} + \theta_h\II\left\{\mu^\star_b=t_h\right\}\right)
=k. 
\]
When $t_h=t'_h$,
\[
\theta'_h = \frac{R_{\pi'_h}/m-\sum_{b\in[B]}\rho^\star_b\II\left\{\mu^\star_b>t_h\right\}}{\sum_{b\in[B]}\rho^\star_b\II\left\{\mu^\star_b=t_h\right\}}<\theta_h. 
\]
Thus
\begin{align*}
m\sum_{b\in[B]}\rho^\star_b\mu^\star_b\left(\II\left\{\mu^\star_b>t'_h\right\} + \theta'_h\II\left\{\mu^\star_b=t'_h\right\}\right)&=m\sum_{b\in[B]}\rho^\star_b\mu^\star_b\left(\II\left\{\mu^\star_b>t_h\right\} + \theta'_h\II\left\{\mu^\star_b=t_h\right\}\right)\\
&<m\sum_{b\in[B]}\rho^\star_b\mu^\star_b\left(\II\left\{\mu^\star_b>t_h\right\} + \theta_h\II\left\{\mu^\star_b=t_h\right\}\right)\\
&=k. 
\end{align*}
This shows that no policy in $\Pi_h$ can achieve smaller objective than $\pi^\star_h$, while satisfying the constraint. So $\pi^\star_h$ is a solution to the minimization problem among $\Pi_h$. 

\subsection{Proof of Corollary~\ref{cor:dominance}}
\label{app:cor-dominance}
By the definition of $\Pi_h$, it is easy to see that the policy space of using $h'$ is a subset of that of $h$ $\Pi_{h'}\subseteq\Pi_h$, the corollary directly follows from Theorem~\ref{thm:perfectly-calibrated-classifier-marginal}. 

\subsection{Proof of Theorem~\ref{thm:optimality_threshold_decision_rules}}
\label{app:thm-optimality_threshold_decision_rules}

The proof constructs a policy $\pi_{f,t}$ for any policy $\pi_f$ such that the two inequalities in the theorem hold. For any policy $\pi_f$, let
\[
k_{\pi_f}\coloneqq\EE_{(\Xb,\Yb)\sim P, \Sbb \sim \pi_f}\left[\sum_{i\in[m]}S_iY_i\right]
\] 
be the expected number of qualified candidates selected by $\pi_{f}$, $\mu_{f}(\cdot)=\Pr\rbr{Y=1\given f(X)=\cdot}$ denote the conditional probability that a candidate is qualified given a prediction quality score from $f$. We construct the the threshold in policy $\pi_{f,t}$ as
\[
t = \sup\left\{u\in[0,1]:m\EE_{f(X)\sim P_{f(X)}}\left[\mu_f(f(X)) \II\cbr{f(X) \geq u}\right] \geq k_{\pi_f}\right\}. 
\]
The expected number of qualified candidates using $\pi_{f,t}$ is
\begin{align*}
    \EE_{(\Xb,\Yb)\sim P, \Sbb \sim \pi_{f,t}}\left[\sum_{i\in[m]}Y_iS_i\right]
    &=\EE_{\{(f(X_i),Y_i)\}_{i\in[m]}\sim P_{f(X),Y}^m,\Sbb\sim \pi_{f,t}\left(\{f(X_i)\}_{i\in[m]}\right)}\left[\sum_{i\in[m]}Y_i S_i\right]\\
    &=\EE_{\{f(X_i)\}_{i\in[m]}\sim P_{f(X)}^m}\left[\sum_{i\in[m]}\mu_f(f(X_i)) \II\cbr{f(X_i) \geq t}\right]\\
    &=m\EE_{f(X)\sim P_{f(X)}}\left[\mu_f(f(X)) \II\cbr{f(X) \geq t}\right]\\
    &= k_{\pi_f}. 
\end{align*}
This shows that the expected number of qualified candidates selected by $\pi_{f,t}$ is no smaller than $\pi_f$. Now we only need to show the expected size of the shortlist using $\pi_{f,t}$ is no larger than that of using $\pi_f$. 
The expected size of the shortlist using $\pi_{f,t}$ is
\begin{align*}
\EE_{\Xb \sim P, \Sbb\sim\pi_{f,t}}\left[\sum_{i\in[m]}S_i\right]
= \EE_{\cbr{f(X_i)}_{i\in[m]} \sim P_{f(X)}^m}\sbr{\sum_{i\in[m]}\II\cbr{f(X_i) \geq t}}
&= m\EE_{f(X) \sim P_{f(X)}}\sbr{\II\cbr{f(X) \geq t}}\\
&=m\Pr\rbr{f(X)\geq t}. 
\end{align*}

We will show that $\pi_f$ achieves no smaller expected size of the shortlists by using contradiction. 
Suppose $\pi_f$ achieves smaller expected size of the shortlist than $\pi_{f,t}$
\begin{align*}
\EE_{\Xb \sim P, \Sbb \sim \pi_f}\left[\sum_{i\in[m]}S_i\right]\coloneqq R_{\pi_f}< & m \Pr\rbr{f(X) \geq t}, 
\end{align*}
we will show that under this assumption, we will arrive at a contradiction that the expected number of qualified candidates selected by $\pi_{f}$ is smaller than $k_{\pi_f}$. 
Let
\[
t' = \sup\left\{u\in[0,1]: m \Pr\rbr{f(X) \geq u}  \geq R_{\pi_f}\right\}. 
\]
Note that $t < t'$ since
\[
m\Pr\rbr{f(X) \geq t} > R_{\pi_f}. 
\]
Then, the expected number of qualified candidates using $\pi_f$ can be bounded as 
\begin{align*}
    \EE_{(\Xb, \Yb)\sim P, \Sbb \sim \pi_f}\left[\sum_{i\in[m]}Y_iS_i\right]&= \EE_{\cbr{f(X_i)}_{i\in[m]} \sim P_{f(X)}^m, \Sbb\sim\pi_f\rbr{\cbr{f(X_i)}_{i\in[m]}}}\sbr{\sum_{i\in[m]}\mu_{f}(f(X_i)) S_i}\\
    &\leq \EE_{\cbr{f(X_i)}_{i\in[m]} \sim P_{f(X)}^m, \Sbb\sim\pi_{f,t'}\rbr{\cbr{f(X_i)}_{i\in[m]}}}\sbr{\sum_{i\in[m]}\mu_{f}(f(X_i)) S_i}\\
    &=m\EE_{f(X)\sim P_{f(X)}}\sbr{\mu_f(f(X))\II\cbr{f(X) \geq t'}}\\
    & < m\EE_{f(X)\sim P_{f(X)}}\sbr{\mu_f(f(X))\II\cbr{f(X) \geq t}}\\
    &= k_{\pi_f}. 
\end{align*}
The inequality is by monotone property of $f$. This arrives at a contradiction with the definition that the expected size of shortlist using $\pi_f$ is $k_{\pi_f}$. So, we get 
\[
\EE_{\Xb \sim P, \Sbb \sim \pi_f}\sbr{\sum_{i\in[m]}S_i} \geq \EE_{\Xb \sim P, \Sbb \sim \pi_{f,t}}\sbr{\sum_{i\in[m]}S_i}, 
\]
which concludes the proof. 



\subsection{Proof of Proposition~\ref{prop:bounds_errors_calibration}}
\label{app:prop-bounds_errors_calibration}
Before introducing the proof of the proposition, we first introduce the DKWM inequality~\cite{dvoretzky1956asymptotic,massart1990tight} as a lemma. \begin{lemma}[DKWM inequality~\cite{dvoretzky1956asymptotic,massart1990tight}]
\label{lemma:DKWM_inequality}
Given a natural number $n$, let $Z_1$, $Z_2$,\dots, $Z_n$ be real-valued independent and identically distributed random variables with cumulative distribution function $F(\cdot)$. Let $F_n$ denote the associated empirical distribution function defined by
\[
F_n(z) = \frac{1}{n}\sum_{i\in[n]}\II\left\{Z_i\leq z\right\}\quad z\in\RR. 
\]
For any $\alpha\in(0,1)$, with probability $1-\alpha$
\[
\sup_{z\in\RR}\left|F(x) - F_n(z)\right|\leq \sqrt{\frac{\ln(2/\alpha)}{2n}}. 
\]
\end{lemma} 

Equipped with this lemma, we provide the proof for Proposition~\ref{prop:bounds_errors_calibration} as follows. 

Let
\[
Z_i^\delta = - Y^c_if(X^c_i)\quad \forall i\in[n]. 
\]
They are real-valued independent and identically distributed random variables. Let $F^\delta(\cdot)$ denote the cumulative distribution function of $Z^\delta$ and
\[
F^\delta_n(z)=\frac{1}{n}\sum_{i\in[n]}\II\left\{ - Y_i^cf(X^c_i)\leq z\right\}\quad\forall z\in\RR. 
\]
Applying the DKWM inequality, we can get that for any $\alpha\in(0,1)$, with probability at least $1-\alpha$, for any $z\in\RR$, 
\[
\left|F^\delta_n(z)-F^\delta(z)\right|\leq \sqrt{\frac{\ln(2/\alpha)}{2n}}.
\]
Note that
\[
F_n^\delta(t) = -\hat{\delta}_{t,1},
\]
and 
\[
F^\delta(t) = -\delta_{t,1},
\]
which concludes the proof. 

\subsection{Proof of Corollary~\ref{cor:bounds_on_objectives}}
\label{app:cor-bounds_on_objectives}
When the event in Proposition~\ref{prop:bounds_errors_calibration} holds, 
\begin{align*}
\EE_{(\Xb,\Yb)\sim P, \Sbb\sim \pi_{f,t}}\left[\sum_{i\in[m]}S_iY_i\right]&=\EE_{\cbr{(f(X_i),Y_i)}_{i\in[m]}\sim P_{f(X),Y}^m, \Sbb\sim \pi_{f,t}\left(\left\{f(X_i)\right\}_{i\in[m]}\right)}\left[\sum_{i\in[m]}S_i Y_i\right]\\
&=\EE_{\cbr{(f(X_i), Y_i)}_{i\in[m]}\sim P_{f(X),Y}^m}\left[\sum_{i\in[m]}Y_i\II\cbr{f(X_i)\geq t}\right]\\
&=m\EE_{(f(X), Y)\sim P_{f(X),Y}}\left[Y\II\cbr{f(X)\geq t}\right]\\
&=m\delta_{t,1}. 
\end{align*}
By Proposition~\ref{prop:bounds_errors_calibration}, this corollary holds, which concludes the proof. 

\subsection{Proof of Theorem~\ref{thm:calibration_from_data_threshold}}
\label{app:thm-calibration_from_data_threshold}
We can see that
\[
\EE_{\Xb\sim P, \Sbb\sim \pi_{f,t}}\left[\sum_{i\in[m]}S_i\right]
\]
is a non-increasing function of $t$. So to solve the constraint minimization problem, we only need to select the largest $t$ that satisfies the constraint, which is Eq.~\ref{eq:calibration_from_data_threshold}. This concludes the proof. 

\subsection{Proof of Proposition~\ref{prop:gap_calibration_from_data}}
\label{app:prop_gap_calibration_from_data}
For the optimal policy $\pi_{f,t^\star_f}$, the constraint is active (otherwise we can randomly select the candidates selected by $\pi_{f,t^\star_f}$ to derive a better policy, which contradicts that $\pi_{f,t^\star_f}$ is optimal)
\[
\EE_{(\Xb,\Yb) \sim P, \Sbb\sim\pi_{f,t^\star_f}}\sbr{\sum_{i\in[m]}S_iY_i} = k. 
\]
Since the distribution of $f(X)$ is nonatomic, all the candidates have different prediction scores almost surely. Thus, by the definition of $\hat{t}_f$, we know that 
\[
\hat{\delta}_{\hat{t}_f,1} - \epsilon(\alpha,n)<k/m + 1/n
\]
almost surely. 
By Proposition~\ref{prop:bounds_errors_calibration} and the proof in Corollary~\ref{cor:bounds_on_objectives}, we have that for any $\alpha\in(0,1)$, with probability at least $1-\alpha$, 
\[
\EE_{(\Xb,\Yb) \sim P, \Sbb\sim\pi_{f,\hat{t}_f}}\sbr{\sum_{i\in[m]}Y_i S_i}= m\delta_{\hat{t}_f,1} \leq m \rbr{\hat{\delta}_{\hat{t}_f,1} + \epsilon(\alpha,n)}<k + 2m\epsilon(\alpha,n) + m/n. 
\]
This concludes the proof. 

\subsection{Proof of Proposition~\ref{prop:PAC_individual_guarantee}}
\label{app:prop_PAC_individal_guarantee}

Let $Q_{i} = S_iY_i$  for all $i\in[m]$ be the random variables of whether a candidate $i$ is qualified and selected in the pool of candidates. Note that 
\[
\EE_{(\Xb,\Yb)\sim P, \Sbb\sim\pi_{f,\hat{t}_f}}[Q_{i}] = \delta_{\hat{t}_f,1}. 
\]

By Theorem~\ref{thm:calibration_from_data_threshold}, we know that for any $\alpha_1\in(0,1)$ with probability at least $1-\alpha_1$,
\[
\EE_{(\Xb,\Yb)\sim P, \Sbb\sim\pi_{f,\hat{t}_f}}\sbr{\sum_{i\in[m]}Q_i}=m\delta_{\hat{t}_f,1}\geq k. 
\]

By applying Bernstein's inequality to these $m$ independent and identically distributed random variables, we have that, for any $\alpha_2\in(0,1)$, with probability at least $1-\alpha_2$
\[
\sum_{i\in[m]}Q_i\geq m\delta_{\hat{t}_f,1} -\frac{1}{3}\ln(1/\alpha_2) -  \sqrt{1/9\ln^2(1/\alpha_2) + 2m\delta_{\hat{t}_f,1}\ln(1/\alpha_2)}. 
\]

The right hand side of the above inequality is an increasing function of $m\delta_{\hat{t}_f,1}$ when $m\delta_{\hat{t}_f,1}\geq 4/9\ln(1/\alpha_2)$. When $k\geq 4/9\ln(1/\alpha_2)$, we can apply the union bound to the above two events to get that, with probability at least $1-\alpha_1 - \alpha_2$,
\[
\sum_{i\in[m]}Q_i \geq m\delta_{\hat{t}_f,1} - \frac{1}{3}\ln(1/\alpha_2) -  \sqrt{1/9\ln^2(1/\alpha_2) + 2m\delta_{\hat{t}_f,1}\ln(1/\alpha_2)}
\geq k - \frac{1}{3}\ln(1/\alpha_2) - \frac{1}{3}\sqrt{\ln^2(1/\alpha_2) + 18k\ln(1/\alpha_2)}.
\] 
This concludes the proof. 

\subsection{Proof of Proposition~\ref{prop:pool-independent-policy-representative}}
\label{app:prop-pool-independent-policy-representative}
We prove the proposition for $f$ that has a discrete range, and the proof for classifiers with continuous range can be derived similarly. 
For any $\pi_f$, we construct a pool-independent policy $\pi'_f$ by constructing $p_{\pi'_f}\in\range(f)\rightarrow [0,1]$ as
\[
p_{\pi'_f}(v) = \frac{\EE_{M \sim P_{M}, \cbr{f(X_i)}_{i\in[M]}\sim P_{f(X)}^M,\Sbb \sim \pi_f}\sbr{\sum_{i\in[M]}\II\cbr{f(X_i) = v} S_i}}{\EE\sbr{M}\EE_{f(X)\sim P_{f(X)}}\cbr{\II\cbr{f(X) = v}}}. 
\]
Let $\pi_f'$ be the policy using $p_{\pi'_f}$. The expected number of qualified candidates for policy $\pi'_f$ equals that of $\pi_f$
\begin{align*}
&\EE_{(M,\Xb,\Yb)\sim P, \Sbb\sim\pi'_f}\sbr{\sum_{i\in[M]}Y_iS_i}\\ 
&= \EE_{M\sim P_{M}, \cbr{f(X_i)}_{i\in[M]}\sim P_{f(X)}^M, \Sbb\sim\pi'_f\rbr{\cbr{f(X_i)}_{i\in[M]}}}\sbr{\sum_{i\in[M]}S_i \Pr\rbr{Y=1 \given f(X_i)}}\\
&=\EE_{M\sim P_{M}, \cbr{f(X_i)}_{i\in[M]} \sim P^M_{f(X)}}\sbr{\sum_{i\in[M]} p_{\pi'_f}(f(X_i))\Pr\rbr{Y=1 \given f(X_i)} }\\
&= \EE\sbr{M} \EE_{v\sim P_{f(X)}}\sbr{p_{\pi'_f}(v)\Pr\rbr{Y=1\given v}}\\
&= \EE_{v\sim P_{f(X)}}\sbr{\frac{\EE_{M \sim P_{M}, \cbr{f(X_i)}_{i\in[M]}\sim P_{f(X)}^M,\Sbb \sim \pi_f}\sbr{\sum_{i\in[M]}\II\cbr{f(X_i) = v} S_i}\Pr\rbr{Y=1\given v}}{\EE_{f(X)\sim P_{f(X)}}\cbr{\II\cbr{f(X) = v}}}}\\
&=\EE_{M\sim P_{M}, \cbr{f(X_i)}_{i\in[M]}\sim P_{f(X)}^M, \Sbb\sim\pi_f\rbr{\cbr{f(X_i)}_{i\in[M]}}}\sbr{\sum_{i\in[M]}S_i \Pr\rbr{Y=1 \given f(X)}}\\
&=\EE_{(M,\Xb,\Yb)\sim P, \Sbb\sim\pi_f}\sbr{\sum_{i\in[M]}Y_iS_i}, 
\end{align*}

and the expected size of the shortlist equals that of $\pi_f$

\begin{align*}
\EE_{(M,\Xb,\Yb)\sim P, \Sbb\sim\pi'_f}\sbr{\sum_{i\in[M]}Y_iS_i} 
&= \EE_{M\sim P_{M}, \cbr{f(X_i)}_{i\in[M]}\sim P_{f(X)}^M, \Sbb\sim\pi'_f\rbr{\cbr{f(X_i)}_{i\in[M]}}}\sbr{\sum_{i\in[M]}S_i}\\
&=\EE_{M\sim P_{M}, \cbr{f(X_i)}_{i\in[M]} \sim P^M_{f(X)}}\sbr{\sum_{i\in[M]} p_{\pi'_f}(f(X_i))}\\
&= \EE\sbr{M} \EE_{v\sim P_{f(X)}}\sbr{p_{\pi'_f}(v)}\\
&= \EE_{v\sim P_{f(X)}}\sbr{\frac{\EE_{M \sim P_{M}, \cbr{f(X_i)}_{i\in[M]}\sim P_{f(X)}^M,\Sbb \sim \pi_f}\sbr{\sum_{i\in[M]}\II\cbr{f(X_i) = v} S_i}}{\EE_{f(X)\sim P_{f(X)}}\cbr{\II\cbr{f(X) = v}}}}\\
&=\EE_{M\sim P_{M}, \cbr{f(X_i)}_{i\in[M]}\sim P_{f(X)}^M, \Sbb\sim\pi_f\rbr{\cbr{f(X_i)}_{i\in[M]}}}\sbr{\sum_{i\in[M]}S_i}\\
&=\EE_{(M,\Xb,\Yb)\sim P, \Sbb\sim\pi_f}\sbr{\sum_{i\in[M]}Y_iS_i}. 
\end{align*}

\subsection{Proof of Proposition~\ref{prop:calibration-errors-multiple-bins}}

From the proof of Proposition~\ref{prop:bounds_errors_calibration}, we know that with probability at least $1-\alpha$
\[
\left|\EE_{(X,Y)\sim P_{X,Y}}\sbr{Y\II\cbr{f(X) \geq t}} - \frac{1}{n}\sum_{i\in[n]} Y_i^c\II\cbr{f(X_i^c)\geq t}\right| \leq \epsilon(\alpha,n) 
\]
for all $t\in[0,1]$. So, with probability at least $1-\alpha$, for any $0\leq t_a< t_b\leq 1$, 

\begin{multline*}
\left|\EE_{(X,Y)\sim P_{X,Y}}\sbr{Y\II\cbr{t_a \leq f(X) \leq t_b}} - \frac{1}{n}\sum_{i\in[n]} Y_i^c\II\cbr{t_a\leq f(X_i^c)\leq t_b}\right|\\
\leq \left|\EE_{(X,Y)\sim P_{X,Y}}\sbr{Y\II\cbr{f(x)\geq t_a}} - \frac{1}{n}\sum_{i\in[n]} Y_i^c\II\cbr{f(X_i^c)\geq t_a}\right|\\
+\left|\EE_{(X,Y)\sim P_{X,Y}}\sbr{Y\II\cbr{f(X) \geq t_b}} - \frac{1}{n}\sum_{i\in[n]} Y_i^c\II\cbr{f(X_i^c)\geq t_b}\right|
\leq 2\epsilon(\alpha,n). 
\end{multline*}
This concludes the proof. 
\end{document}